\documentclass[11pt,a4paper]{article}

\usepackage{graphicx}

\usepackage{palatino,epsfig,latexsym}

\usepackage{geometry}
\usepackage{pdflscape}
\usepackage{amsfonts}
\usepackage{amssymb}
\usepackage{amsmath}
\usepackage{rotating}
\usepackage{subfigure}
\usepackage{algorithm,xcolor}
\usepackage{algorithmic}
\usepackage[utf8]{inputenc}
\usepackage[T1]{fontenc}

\usepackage{multirow,booktabs}
\usepackage{tikz}
\usetikzlibrary{decorations.pathmorphing}
\usepackage{hyperref}

\usepackage{array}

\usepackage{hyphenat}
\definecolor{officegreen}{rgb}{0.0, 0.5, 0.0}

\begin{document}

\title{An Extensive Experimental Evaluation of Automated Machine Learning Methods for Recommending Classification Algorithms (Extended Version)}



\author{M\'arcio P.~Basgalupp \and
        Rodrigo C.~Barros \and
        Alex G.~C.~de~S\'a \and
        Gisele Pappa \and
        Rafael G.~Mantovani \and
        Andr\'e C.~P.~L.~F.~de~Carvalho \and
        Alex A. Freitas
}

\date{Received: date / Accepted: date}

\maketitle

\begin{abstract}

This paper presents an experimental comparison among four Automated Machine Learning (AutoML) methods for recommending the best classification algorithm for a given input dataset. Three of these methods are based on Evolutionary Algorithms (EAs), and the other is Auto-WEKA, a well-known AutoML method based on the Combined Algorithm Selection and Hyper-parameter optimisation (CASH) approach. The EA-based methods build classification algorithms from a single machine learning paradigm: either decision-tree induction, rule induction, or Bayesian network classification. Auto-WEKA combines algorithm selection and hyper-parameter optimisation to recommend classification algorithms from multiple paradigms. We performed controlled experiments where these four AutoML methods were given the same runtime limit for different values of this limit. In general, the difference in predictive accuracy of the three best AutoML methods was not statistically significant. However, the EA evolving decision-tree induction algorithms has the advantage of producing algorithms that generate interpretable classification models and that are more scalable to large datasets, by comparison with many algorithms from other learning paradigms that can be recommended by Auto-WEKA. We also observed that Auto-WEKA has shown meta-overfitting, a form of overfitting at the meta-learning level, rather than at the base-learning level.
\end{abstract}



\section{Introduction}
\label{sec:introduction}

Classification is one of the main machine learning tasks and, hence, there is a large variety of classification algorithms available \cite{Witten2016, Zaki2020}. However, in most real-world applications, the choice of classification algorithm for a new dataset or application domain is still mainly an ad-hoc decision.

In this context, the use of meta-learning for algorithm recommendation is a very important research area with seminal work dating back more than $20$ years, which includes the StatLog \cite{Michie1994} and METAL \cite{METAL} projects.
Meta-learning can be defined as learning how to learn, which involves learning, from previous experience, what is the best machine learning algorithm (and its best hyper-parameter setting) for a given dataset \cite{Brazdil2008, Vanschoren2018}.
Meta-learning systems for algorithm recommendation can be divided into two broad groups, namely: (a) systems that perform algorithm selection based on meta-features ~\cite{Brazdil2008}, which is the most investigated type; and (b) systems that search for the best possible classification algorithm in a given algorithm space \cite{Thornton2013autoweka}.

Meta-feature-based meta-learning for algorithm selection and recommendation consists of two basic steps~\cite{Brazdil2008}. First, the creation of a meta-training set where each meta-instance represents a dataset, meta-features represent dataset properties, and each meta-class represents a (base level) learning algorithm. Second, the induction of a meta-classification model by a (meta) classification algorithm over the meta-training set, thus allowing the recommendation of algorithm(s) for a novel dataset (not included in the meta-training set). A key issue is the design of a good set of meta-features, with enough predictive power to support an accurate recommendation of the best learning algorithm. Extensive research in this topic has produced a large variety of meta-features \cite{Brazdil2008,Ho2002a,Ho2006}, but the issue of finding a set of meta-features with very good predictive power is still an open and difficult problem. 

A limitation of meta-feature-based meta-learning research is that usually a small number of candidate classification algorithms are considered as meta-classes. This is because in general, the larger the number of candidate classification algorithms used as meta-classes, the more difficult it would be for the meta-classification algorithm to accurately predict all meta-classes. In addition, it is difficult to produce large meta-datasets for meta-learning, since in order to compute the meta-class of each meta-instance we need to run all candidate classification algorithms on all datasets (one for each meta-instance). 

These difficulties have motivated research on the second type of meta-learning for algorithm recommendation, meta-learning systems using search or optimisation methods to indicate the best classification algorithm for a given target dataset, in a given algorithm space \cite{Pappa2009,AT2012,Thornton2013autoweka,Pappa2014,autoWeka2.0,Barros2015book,LC2015}. 
This work focuses mainly on this type of meta-learning systems, which is a type of Automated Machine Learning (AutoML) \cite{Hutter2019}, since such systems effectively automate the process of selecting the best algorithm and its hyper-parameters for the input dataset.

This AutoML approach bypasses the need for designing meta-features and it can, in principle, consider a substantially larger number of candidate classification algorithms and hyper-parameters than meta-feature-based meta-learning systems. Note that although this approach does not explicitly use a learning algorithm at the meta-level, some methods following this AutoML approach (like some methods evaluated in this work) perform a form of meta-learning because the search is performed in the space of candidate learning algorithms and is guided by an evaluation function based on the accuracy of learning algorithms at the base level. Therefore, the search method at the meta-level is implicitly learning from the results of base-level learning algorithms. Note, however, that this kind of meta-learning of course does not occur in the case of simple and popular methods for algorithm selection and parameter configuration, like random search and grid search, which do not perform any learning by themselves.

In this context, the main contribution of this paper is to present an extensive empirical comparison of the predictive performance of four sophisticated AutoML methods for the recommendation of classification algorithms. One of these methods, Auto-WEKA~\cite{Thornton2013autoweka,autoWeka2.0}, performs algorithm selection and hyper-parameter configuration by considering all candidate classification algorithms available in the well-known WEKA data mining tool, which includes algorithms based on several different types of knowledge (or model) representations -- e.g., decision trees, if-then classification rules, Bayesian network classifiers, neural networks, support vector machines, etc. 
The other three methods are based on evolutionary algorithms (EAs). Unlike Auto-WEKA, each of the three EAs focuses on a search space containing classification algorithms based on a single type of knowledge representation. More precisely, the EAs evolve rule induction algorithms \cite{Pappa2009}, decision-tree induction algorithms \cite{Barros2015book}, and Bayesian network classification algorithms \cite{Sa2013}. Hence, the EAs produce a narrower diversity of classification algorithms in terms of knowledge representation. However, within its specialized knowledge representation, an EA can have more flexibility (or autonomy) to construct new classification algorithms, rather than just optimising the configuration of hyper-parameters for an existing classification algorithm, as discussed later. 

There are also other recently proposed EAs for related AutoML tasks. In particular, the EAs proposed in~\cite{deSa2017, Kren2017, Olson2016} try to optimize an entire machine learning pipeline for a given dataset, including the choice of data preprocessing methods (like feature scaling operators and feature selection methods) and classification algorithm. By contrast, we focus on using EAs that recommend only classification algorithms. In addition, in~\cite{NyathiP17} an EA is proposed to automatically evolve another type of EA (genetic programming) for classification. By contrast, the EAs used here automatically evolve more conventional (non-evolutionary) types of classification algorithms, as mentioned earlier.

Controlled experiemnts were performed, where the four previous AutoML methods (the three EAs and Auto-WEKA) had the same runtime limit for different values of this limit. In general, the difference in predictive accuracy of the three best AutoML methods was not statistically significant, but Auto-WEKA showed meta-overfitting, a form of overfitting at the meta-learning level, due to evaluating many different (base-level) classification algorithms during its search for the best algorithm. This is in contrast to the standard overfitting at the base level, due to the evaluating many different models built by the same classification algorithm. In addition, the EA evolving decision-tree induction algorithms have the advantage of producing algorithms that generate interpretable classification models and that are more scalable to large datasets, by comparison with many algorithms from other learning paradigms that can be recommended by Auto-WEKA. Furthermore, an analysis of the different types of classification algorithms recommended by Auto-WEKA shows that overall decision-tree and ensemble algorithms were the most frequently recommended types of algorithms, whilst rule induction algorithms were the least recommended type.

The remainder of this paper is organised as follows. Section~\ref{sec:methods} reviews the background on AutoML methods for classification-algorithm recommendation, focusing on the four previously mentioned AutoML methods. Section~\ref{sec:experiments} describes the methodology adopted in this study for executing the experimental analyses, whose extensive results are presented in Section~\ref{sec:results}. Finally, the main conclusions and future work suggestions are presented in Section~\ref{sec:conclusions}.


\section{AutoML Methods for Classification-Algorithm Recommendation}
\label{sec:methods}

This section reviews the main concepts underlying several AutoML methods for automatic recommendation of the best classification algorithm for a given input dataset. It mainly covers the four AutoML methods evaluated in this work, Auto-WEKA and three EAs, as mentioned earlier. 
Its last subsection briefly reviews related work on other evolutionary AutoML methods.


\subsection{Auto-WEKA and the CASH Problem}

Initial work on meta-learning focused on selecting the best classification algorithm(s) for a given dataset, explicitly or implicitly assuming a default configuration (hyper-parameter settings) for the candidate algorithms. However, given that the success of a classification algorithm strongly depends on its hyper-parameter settings, more recent work has focused on the so called Combined Algorithm Selection and Hyper-parameter (CASH) optimisation problem \cite{Thornton2013autoweka}. In this section, we review the AutoML methods evaluated in this work that address the CASH problem by considering, as candidate algorithms to be recommended, classification algorithms from multiple knowledge (model) representations, like decision trees, IF-THEN classification rules, probabilistic graphical models, neural networks, ensembles, etc.

In this context, an advanced and well-known system designed for the CASH problem is Auto-WEKA \cite{Thornton2013autoweka,autoWeka2.0}, whose search-space includes all classification algorithms available in \texttt{Weka} \cite{Hall2009} with their corresponding candidate hyper-parameter settings. 

In order to search the space of candidate algorithms and their hyper-parameter settings, Auto-WEKA uses a stochastic search method, named Sequential Model-Based Optimisation (SMBO), and a loss function to measure classification error. The goal is to find the classification algorithm and its corresponding hyper-parameter settings that minimise the value of the loss function for the target dataset. SMBO essentially works as follows. First, the CASH problem is formulated as a hierarchical hyper-parameter search-space where there is a new root-level hyper-parameter that selects between algorithms. Hence, a candidate solution is an algorithm selected at the root level and its hyper-parameters selected at lower levels. As shown in Algorithm~\ref{alg:autoweka}, SMBO initially builds a model ($M_L$, line 1) representing the dependency of the loss function on the candidate hyper-parameter settings. Next, it iteratively uses the model to generate a promising candidate hyper-parameter setting ($\lambda$, line 3), evaluates the setting (lines 4-5), and updates the model according to the evaluation (line 6). SMBO is flexible enough to be able to be used with different algorithms for building the dependency model, with random forests being used in \cite{Thornton2013autoweka,autoWeka2.0}.

\begin{algorithm}
\scriptsize
\caption{Pseudo-code of SMBO. Adapted from \cite{Thornton2013autoweka}.}
\begin{algorithmic}[1]
\STATE Initialise model $M_L$; $H = \emptyset$
\WHILE {time budget has not been exceeded}
    \STATE $\lambda = $ candidate configuration from $M_L$
    \STATE compute $c = L(A_{\lambda},D^{(i)}_{train},D^{(i)}_{valid})$
    \STATE $H = H \cup \{(\lambda,c) \}$
    \STATE Update $M_L$ given $H$
\ENDWHILE
\RETURN $\lambda$ from $H$ with minimal $c$
\end{algorithmic}
\label{alg:autoweka}
\end{algorithm}

The approach used by Auto-WEKA was also extended to produce another system for solving the CASH problem, namely Auto-sklearn \cite{Feurer2015a}, which uses the scikit-learn machine learning library \cite{scikit-learn} rather than \texttt{Weka}. Auto-sklearn extends Auto-WEKA's approach in two ways. First, it uses an ensemble of the classification models generated by the SMBO search method, instead of just one model like in Auto-WEKA. Second, it uses meta-features-based meta-learning to find good classification algorithm configurations (see \cite{Feurer2015a, Feurer2015c} for details of these two extensions). In addition, meta-features-based meta-learning has been recently used to initialise the SMBO's search for the optimal solution to the CASH problem \cite{Feurer2015b}. It should be noted that the aforementioned systems, although very advanced, are limited to find a combination of algorithm and hyper-parameter settings among existing combinations in the base machine learning toolkit being used (\texttt{Weka} or scikit-learn). They do not have enough autonomy for constructing a new classification algorithm, which can be done in some cases by the EA-based meta-learning methods discussed in the next section.

\subsection{EA-based AutoML Methods}

Each of the Evolutionary Algorithm-based (EA-based) AutoML methods evaluated in this work explores a search space with classification algorithms from a different knowledge (model) representation, namely: rule induction \cite{Pappa2009}, decision-tree induction  \cite{Barros2012a}, or Bayesian network classifiers \cite{Sa2013}. 

EAs are search methods based on the natural selection principle \cite{Eiben2015}. They have been extensively used for evolving classification models in machine learning \cite{Freitas2008b,Barros2012a}. In this work, however, the EAs evolve full classification algorithms rather than classification models. In EA terminology, the EAs used in this work are hyper-heuristic search methods, which perform a search in the space of candidate classification $algorithms$ \cite{Pappa2014}; whilst EAs that perform a search in the space of classification $models$ are conventional meta-heuristic search methods.

The three EAs receive as input a high-level pseudo-code with the main algorithmic components to be used to create classification algorithms from a target algorithm type. For instance, if the target is rule induction algorithms, the components include a rule search method, a rule evaluation criterion, etc. Each component can be instantiated in different ways, e.g., confidence or information gain can be used to instantiate the rule evaluation component. Given an input dataset, an EA searches for the best combination of algorithmic components based on an evaluation function (called fitness function in EAs). Thus, the EA's output is a classification algorithm of the target type.

Note that the EAs can sometimes generate a new classification algorithm which works in a way different from all current (manually-designed) classification algorithms. This is because the EAs can combine the prespecified algorithmic components in novel ways, not explored by human algorithm designers yet.

As an example of algorithm construction, let us consider the EA for evolving decision-tree algorithms. That EA’s algorithmic components include, among other types of components, 15 different split criteria and 5 tree-pruning methods. A manually-designed decision-tree algorithm like J48 (WEKA’s version of C4.5) or CART offers just a subset of these split criteria and pruning methods. Hence, when Auto-WEKA configures a decision-tree algorithm, it first chooses exactly which algorithm will be configured, say J48 or CART, and then it considers only the split criteria and tree pruning methods/hyper-parameters available in WEKA for the chosen algorithm. It cannot combine, e.g., the information gain ratio used by J48 with the cost-complexity pruning used by CART. By contrast, the EA can construct a new decision-tree induction algorithm with any combination of split criteria and tree pruning method/hyper-parameters (as well as any combination of other specific components), regardless of whether or not the chosen combination of components occurs in a current manually-designed decision-tree algorithm.

Algorithm~\ref{alg:hh} shows the high-level pseudo-code of the three EAs for recommending classification algorithms used in this work. First, they generate a population of candidate solutions (classification algorithms), or individuals, based on the target pseudo-code and sets of components given as input. For a fixed number of iterations (generations) $g$, the classification algorithms represented by the individuals in the initial population $P$ are built and run on the input dataset. The input dataset is divided into meta-training, meta-validation, and meta-test sets. In order to measure the fitness (quality) of an individual, its corresponding classification algorithm is executed over the meta-training set to build a classification model. Afterwards, a given predictive performance measure is used to evaluate the model performance on the meta-validation set, and this measure is used as the fitness of the individual. 

\begin{algorithm}[!htbp]
\scriptsize
\caption{Pseudo-code of evolutionary algorithms for generating classification algorithms.}
\begin{algorithmic}
\STATE BuildTailoredAlgorithm(datasets, generalPseudocode, components, $g$, $s$)
\STATE P = CreatePopulation(generalPseudocode, components)
\STATE count = 0
\STATE BestSet = $\emptyset$
\WHILE {count $< g$}
	\FORALL{ indiv in P}
		\STATE BuildAlgorithm(indiv)
		\STATE RunAlgorithm(indiv,dataset)
	\ENDFOR	
	\STATE TournamentSelection(P)
	\STATE Crossover(P)
	\STATE Mutation(P)
	\STATE count = count + 1
	\IF{count mod s}
	\STATE BestSet = best in P
	\STATE Resample dataset
	\ENDIF
\ENDWHILE
\RETURN best in BestSet according to a predictive performance measure
\end{algorithmic}
\label{alg:hh}
\end{algorithm}

To avoid overfitting, at each $s$ generations, the examples belonging to the meta-training and meta-validation sets are resampled, and the best individual found in that sample is saved in $BestSet$. During the EA run, individuals at different generations may be evaluated with different data. Based on the individuals' fitness values, the best candidate classification algorithms are selected to undergo EA operations such as crossover and mutation, according to user-defined probabilities. At the end of an EA run, the best algorithm output by the EA is chosen as follows. Considering the individuals saved in $BestSet$, a new cross-validation procedure is performed on the training set. All individuals are then executed using the same cross-validation folds, and the best classification algorithm is output. That algorithm is finally evaluated on the meta-test set, which was not seen during the EA run, to compute the final measure of predictive accuracy for the evolved classification algorithm.

All three EAs discussed in this paper follow Algorithm~\ref{alg:hh}, but they vary on how they represent individuals, the types of components used to build classification algorithms (depending on the type of target classification algorithm), and the performance measure used to select the best individuals. All algorithms require user-defined hyper-parameters which include, besides the number of iterations (generations), the number of individuals, the rates of crossover and mutation (operators used to produce new individuals from existing ones), the rate of elitism (i.e. the percentage of individuals from the current generation that are passed unaltered to the next generation), and the number of individuals selected to undergo tournament selection.


\subsubsection{Evolving Rule Induction Algorithms with Grammar-based Genetic Programming}  

\hfill\\
The first EA proposed for generating a full classification algorithm customised to a given input dataset evolves rule induction algorithms (which output IF-THEN classification rules), using a Grammar-based Genetic Programming (GGP) algorithm \cite{Pappa2009}, named GGP-RI (GGP for Rule Induction). GGPs differ from standard EAs as they receive as input a grammar, and all candidate solutions generated must obey the grammar production rules. 

The grammar has production rules specifying how the following components of induction algorithms can be instantiated and combined together into valid algorithms: the decision to generate an unordered rule set or an ordered rule list, different methods to initialize, search, evaluate and prune rules, as well as different loop structures and conditional statements to control the iterative processes of constructing a rule and adding/removing rules to/from a set/list. Each individual is represented by a tree generated by applying the production rules. Each tree is mapped to a rule induction algorithm. The GGP grammar has 26 non-terminals and 83 production rules, and, varying the order in which the production rules are applied, the GGP's search-space has over 2 billion different rule induction algorithms. GGP's fitness function is the F-Measure (the harmonic mean of precision and recall) of a candidate rule induction algorithm in the meta-validation set (as explained earlier).


\subsubsection{Evolving Decision-Tree Induction Algorithms with a Hyper-Heuristic Evolutionary Algorithm}

\hfill\\
A hyper-heuristic EA that generates decision-tree induction algorithms, called HEAD-DT (Hyper-heuristic Evolutionary Algorithm for Automatically Designing Decision-Tree algorithms), is described in \cite{Barros2013ecj,Barros2014a}. Unlike GGP, HEAD-DT is based on a genetic algorithm with linear encoding. An individual (candidate decision-tree induction algorithm) consists of a set of many options to instantiate the following components of decision-tree induction algorithms: the data split procedure used at each node of the tree (i.e., whether performing a binary or multi-way split and which feature evaluation function should be used), the tree expansion stopping criteria, approaches to cope with missing values (in both the training and testing phases), and the tree pruning procedure. For each algorithmic component, an individual specifies both categorical options (e.g., the choice of feature evaluation function, out of 16 predefined functions) and the numerical value of hyper-parameters associated with the chosen options (e.g., a hyper-parameter that controls the degree of pruning for a given pruning method). HEAD-DT's fitness function is the F-Measure of a candidate decision-tree induction algorithm in the meta-validation set, and its search space contains 21,319,200 different decision-tree algorithms. It was applied with success in different application domains, such as gene expression classification \cite{Barros2014a} and rational drug design \cite{Barros2012bmc}. 


\subsubsection{Evolving Bayesian Network Classification Algorithms with a Hyper-Heuristic Evolutionary Algorithm} 

\hfill\\
The EA for generating Bayesian Network Classification (BNC) algorithms is named HHEA-BNC (Hyper-Heuristic Evolutionary Algorithm for creating a BNC algorithm) \cite{Sa2013, Sa2013b}. BNC algorithms usually have two phases \cite{Cheng1999, Daly2011}: (i) network-structure learning; and (ii) parameter learning. In the first phase, the algorithm learns which nodes (features) in the network should be connected to each other. The parameter learning phase, in turn, learns the Conditional Probability Tables (CPTs) for each node of the network (the BNC model). However, learning the parameters of a BNC model is a relatively straightforward procedure when the network structure has been determined. For this reason, HHEA-BNC focuses on the structure learning phase. 
HHEA-BNC encodes candidate BNC algorithms using a dynamic array-like representation, where each position in the array represents a different algorithm component to be instantiated. In order to select and instantiate the components of the BNC algorithm, HHEA-BNC uses a top-down approach, where the first instantiated component of the BNC algorithm being created is the search method, with a choice among $12$ different methods. The search method defines the type of algorithm being generated (na\"ive Bayes, score-based, constraint-based or hybrid) and, consequently, the type of BNC model being created (i.e. tree, graph, or no edges between features, in the case of na\"ive Bayes). Based on this first choice, different BNC algorithms can be generated, including components like scoring metrics, statistical independence tests, maximal number of parents per node, etc. The smallest individual has three components, while the largest has $11$. The search-space of HHEA-BNC has 60,510,000 different candidate BNC algorithms.
HHEA-BNC's fitness function is the F-measure of a candidate BNC algorithm in the meta-validation set.

\subsubsection{Related Work on EA-based AutoML Methods}
\label{related}

We also have identified three evolutionary AutoML methods that try to optimize the entire classification pipeline: 
(i) Tree-based Pipeline Optimization Tool (TPOT) \cite{Olson2016,Olson2016b};
(ii) Genetic Programming for Machine Learning (GP-ML) \cite{Kren2017}; and
(iii) REsilient ClassifIcation Pipeline Evolution (RECIPE) \cite{deSa2017}.
A pipeline is defined as a machine learning workflow that solves the classification task. 
To solve this type of task, a pipeline may contain data preprocessing methods (e.g., feature normalization or feature selection), must have a classification algorithm (e.g., na\"ive Bayes or a support vector machine) and may have a post-processing approach (e.g., voting or stacking).
Therefore, these methods take into account various aspects of machine learning instead of focusing only on the classification algorithm. 
This means that these methods could select and configure a range of different classification-related methods during the evolutionary search,
as they are not centered on just one type of classification algorithm. This basic principle is also followed by Auto-WEKA and Auto-sklearn, which are well-known non-EA-based AutoML methods.
The aforementioned EA-based AutoML methods are discussed in somewhat more detail next.

TPOT is a genetic programming-based method that searches for the most suitable classification pipeline to the input dataset.
It encompasses (part of) the available methods in the scikit-learn library in its search space, and allows 
different ways of combining the data preprocessing methods (in sequence or in parallel) and the classification algorithms (supporting ensemble approaches or not).
Although TPOT has been designed for general classification, it alternatively has a specific version for bioinformatics studies,
named TPOT-MDR \cite{Sohn2017}.
TPOT-MDR includes two new data preprocessing operators that are used in genetic analyses of human diseases: the Multifactor Dimensionality Reduction (MDR)
and the Expert Knowledge Filter (EKF).
Besides, both versions perform multi-objective search using Pareto selection (based on the well-known NSGA-II algorithm) \cite{Deb2002} with 
two objectives: maximizing the predictive accuracy measure of the pipeline and minimizing the pipeline's overall complexity (which is represented by the number of 
pipeline operators).

The main issue when using TPOT is that it can generate classification pipelines that are invalid or arbitrary during its evolutionary process, i.e.,
pipelines that do not solve the classification task itself.
This happens because TPOT does not impose any constraints when combining the ML components to create the pipelines.
For instance, TPOT can create a pipeline without a classification algorithm \cite{Olson2016}.
This, of course, makes the evolutionary process to waste resources as various individuals would not solve the classification task.
This can be considered a significant drawback of TPOT in the context of the classification task.

GP-ML overcomes this limitation by using a strongly typed genetic programming (STGP) method. 
A STGP method restricts the scikit-learn pipelines in such a way that makes them valid from the machine learning point of view.
In addition, GP-ML applies an asynchronous evolutionary algorithm \cite{Scott2016} instead of a generational one.
\cite{Scott2016} observed that asynchronous evolution is biased towards the evaluation of faster pipelines in some parts of the search space.
However, \cite{Kren2017} consider this bias an advantage to the AutoML task, because a faster pipeline is usually preferable to a slower one,
when both present similar predictive accuracy values.

RECIPE follows the same basic principle of GP-ML, i.e., it only allows the generation of valid pipelines during the evolutionary process.
In order to implement this principle, RECIPE defines a grammar which encompasses the classification knowledge in scikit-learn.
Therefore, RECIPE makes use of a grammar-based genetic programming (GGP) \cite{McKay2010} to perform the search for the most
suitable classification pipeline. The grammar prevents the generation of invalid/arbitrary pipelines, and could also speed up the search.

\section{Experimental Methodology}\label{sec:experiments}

The experiments are divided into two parts. The first part compares the results obtained by the EAs with the results obtained by Auto-WEKA \cite{Thornton2013autoweka}, whose search space 
includes all 33 classification algorithms available in WEKA. These experiments used 20 datasets. 

The second part of the experiments compares one of the EAs (HEAD-DT, the EA evolving decision-tree algorithms) against Auto-WEKA, on an extended set of 40 datasets. The main reason for using a smaller number of datasets in the first type of experiment was the very long computation time associated with comparing four methods. HEAD-DT was chosen because, among the two most successful EAs overall (HEAD-DT and HHEA-BNC, as discussed later), HEAD-DT has the advantage of producing decision tree algorithms which are more scalable to larger datasets than the Bayesian network classification algorithms produced by HHEA-BNC. The datasets used in both types of experiments are described next.


\subsection{Datasets} \label{sec:datasets}

The first part of the experiments focus on $20$ challenging datasets, characterised in general (with one exception) by a small number of instances and a large number of attributes. Table~\ref{tab:datasets} summarises their main characteristics, including number of instances, number of numerical and nominal attributes, percentage of missing values, class balance ratio (class bal.) and number of classes. Class bal. is the ratio of the minority class frequency over the majority class frequency -- values closer to 0 (1) indicate datasets with more (less) class distribution imbalance.
The first 12 datasets in this table are bioinformatics datasets, whilst the last 8 ones are text mining datasets.
The first six datasets involve data from the biology of ageing. Datasets CE-T3, SC-T3, DM-T3, and MM-T3 are described in \cite{Cen2015}; whilst datasets DNA-T3 and DNA-T11 are described in \cite{Freitas2011}. Dataset PS-T3 involves post-synaptic proteins~\cite{Pappa2005}. The 5 microarray datasets are publicly-available microarray gene expression datasets, described in~\cite{Souto2008}.
Finally, the  8 text mining datasets were obtained from OpenML~\cite{OpenML2014}. 

\begin{table}[h]
\centering
\caption{Summary of the 20 datasets used in both the first and the second sets of experiments.}
\label{tab:datasets}
\scriptsize
\begin{tabular*}{\textwidth}
	{@{\extracolsep{\fill}}llrrrrrr}
\toprule
Type & Dataset  & \# inst & \# num & \# nom & \% miss & class bal &\# classes \\
\midrule
\multirow{6}{*}{Ageing}
& CE-T3 & 478 &  0 & 764 & 0 & 0.66 & 2 \\
& DM-T3 & 119 &  0 & 586 & 0 & 0.49 & 2 \\
& MM-T3 & 89 & 0 & 887 & 0 & 0.41 & 2 \\
& SC-T3 & 248 & 0 & 698 & 0 & 0.19 & 2 \\
& DNA-T3 & 139 & 3 & 333 & 9 & 0.31 & 2 \\
& DNA-T11 & 135  & 2 & 103 & 26 & 0.32 & 2 \\
\midrule
PS & PS-T3 & 4303 & 2 & 443 & 0 & 0.06 & 2 \\
\midrule
\multirow{5}{*}{Microarray} & chen-2002 & 179 & 85 & 0 & 0  & 0.72 & 2 \\
& chowdary-2006 & 104 & 182 & 0 & 0 & 0.68 & 2 \\
& nutt-2003-v2 & 28 & 1070 & 0 & 0 & 1.00 & 2 \\
& singh-2002 & 102 & 339 & 0 & 0 & 0.96 & 2 \\
& west-2001 & 49 & 1198 & 0 & 0 & 0.96 & 2 \\
\midrule
\multirow{8}{*}{Text} & dbworld-bodies &  64 & 0 &  4702 & 0 & 0.83 & 2 \\ 
& dbworld-bodies-s &  64 & 0 & 3721 & 0 & 0.83 & 2\\ 
& oh0.wc & 1003 & 3182 & 0 & 0 & 0.26 & 10  \\ 
& oh5.wc & 918 & 3012 & 0 & 0 & 0.40 & 10  \\ 
& oh10.wc & 1050 & 3238 & 0 & 0 & 0.32 & 10  \\
& oh15.wc & 913 & 3100 & 0 & 0 & 0.34 & 10  \\ 
& re0.wc & 1504 & 2886 & 0 & 0 & 0.02 & 13 \\ 
& re1.wc & 1657 & 3758  & 0 & 0 & 0.03 & 25  \\  
\bottomrule
\end{tabular*}
\end{table}

Table~\ref{tab:additionalDatasets} summarises the main characteristics for 20 additional datasets which were used only in the final experiments, comparing HEAD-DT and Auto-WEKA. The first 15 datasets used in this Table were used in \cite{Thornton2013autoweka}, whilst the other 5 datasets where used in \cite{Feurer2015c}.

\begin{table}[h]
\centering
\caption{Summary of the 20 datasets used only in the second set of experiments.}
\label{tab:additionalDatasets}
\scriptsize
\begin{tabular*}{\textwidth}
	{@{\extracolsep{\fill}}llrrrrr}
\toprule
Dataset  & \# inst & \# num & \# nom & \% miss & class bal & \# classes \\
\midrule
    abalone & 4177  & 7     & 1     & 0.00  & <0.01  & 28 \\
    car   & 1728  & 0     & 6     & 0.00  & 0.05  & 4 \\
    convex & 58000 & 784   & 0     & 0.00  & 1.00  & 2 \\
    germancredit & 1000  & 7     & 13    & 0.00  & 0.43  & 2 \\
    krvskp & 3196  & 0     & 36    & 0.00  & 0.91  & 2 \\
    madelon & 2600  & 500   & 0     & 0.00  & 1.00  & 2 \\
    mnist & 62000 & 784   & 0     & 0.00  & 0.80  & 10 \\
    mnistrotationbackimagenew & 62000 & 784   & 0     & 0.00  & 0.81  & 10 \\
    secom & 1567  & 590   & 0     & 4.54  & 0.07  & 2 \\
    semeion & 1593  & 256   & 0     & 0.00  & 0.96  & 10 \\
    shuttle & 58000 & 9     & 0     & 0.00  & <0.01  & 7 \\
    waveform & 5000  & 40    & 0     & 0.00  & 0.98  & 3 \\
    winequalitywhite & 4898  & 11    & 0     & 0.00  & 0.00  & 11 \\
    yeast & 1484  & 8     & 0     & 0.00  & 0.01  & 10 \\
    sick & 3772  & 7     & 22    & 5.54  & 0.07  & 2 \\
    splice & 3190  & 0     & 61    & 0.00  & 0.46  & 3 \\
    kropt & 28056 & 0     & 6     & 0.00  & 0.01  & 18 \\
    quake & 2178  & 3     & 0     & 0.00  & 0.80  & 2 \\
    pc4 & 1458  & 37    & 0     & 0.00  & 0.14  & 2 \\
    magicTelescope & 19020 & 10    & 0     & 0.00  & 0.54  & 2 \\
\bottomrule
\end{tabular*}
\end{table}


\subsection{Evaluation Methodology}

The $10$-fold cross-validation technique (10-cv) ~\cite{Witten2016} was used in the experiments.
Since Auto-WEKA and the Evolutionary Algorithms (EAs) are non-deterministic, their results are an average over $5$ executions, generating, for each method,
1000 algorithms.
All results presented in Section~\ref{sec:results} refer to the predictive accuracy of the recommended algorithms in the test sets.

Two predictive accuracy measures are used. First, the Geometric Mean (GMean) of sensitivity ($Sens$) and specificity ($Spec$)~\cite{Japkowicz2011}, defined as \textbf{GMean} = $\sqrt{Sens \times Spec}$. $Sens$ is the proportion of positive instances that were correctly predicted as positive. $Spec$ is the proportion of negative instances that were correctly predicted as negative. These measures were calculated considering each class in turn as the positive class, and then computing the weighted average of these measures, by weighing the classes according to their relative frequency. The GMean measure was also used to evaluate some datasets in \cite{Cen2015}. The second predictive accuracy measure used is the simple classification accuracy measure used by Auto-WEKA to choose the best algorithm for each dataset.

Statistical significance analysis was applied to the experimental results. In the first set of experiments (comparing four methods), we have adopted Dem\v{s}ar's \cite{Demsar2006} recommendation to use the Friedman test with the adjusted statistic $F_F$ \cite{Iman1980} to compare multiple algorithms over multiple datasets, followed by the Nemenyi post-hoc test for pairwise comparisons. In the final experiment comparing only two methods we have used the Wilcoxon test \cite{Wilcoxon1970}. The main advantage of all these statistical tests is that they are non-parametric, so that they do not make the assumption that the data follows the normal distribution (nor assume any other probability distribution, for that matter). All statistical tests were used with the conventional significance level of 0.05.

\subsection{Settings for the Evolutionary Algorithms (EAs) and for Auto-WEKA} \label{sec:EAsparams}

In order to perform a fair comparison, all  EAs were configured with the same hyper-parameters values, listed in Table~\ref{tab:parameters}. 

\begin{table}[h]
\scriptsize
\centering
\caption{Parameter values for the evolutionary algorithms.}
\label{tab:parameters}

\begin{tabular*}{\textwidth}
	{@{\extracolsep{\fill}}lr}
\toprule
Parameter Description & Value \\
\midrule
Number of individuals & 100 \\
Number of generations before changing the validation set & 5 \\
Tournament selection size & 2 \\
Elitism rate & 5\% \\
Crossover rate & 95\% \\
Mutation rate & 5\% \\
\bottomrule
\end{tabular*}
\end{table}

Table~\ref{tab:param-autoweka} shows the hyper-parameter settings for Auto-WEKA based on the options provided by its Experiment Builder~\cite{Thornton2013autoweka}. Note that the 10-cv mentioned in Table~\ref{tab:param-autoweka} is another cross-validation procedure used by Auto-WEKA, but this time over the training set (generated by the outermost 10-cv) to evaluate its candidate solutions regarding their predictive accuracy.

\begin{table}[h]
\centering
\scriptsize
\caption{Hyper-parameter values for all versions of Auto-WEKA.}
\label{tab:param-autoweka}

\begin{tabular*}{\textwidth}
	{@{\extracolsep{\fill}}ll}
\toprule
\multicolumn{1}{c}{Parameter Description} & \multicolumn{1}{c}{Value(s)} \\
\midrule
\rule{0pt}{1ex} 
Instance generator & 10-fold cross-validation, seed = 1,..,5 \\
\rule{0pt}{3ex} 

Evaluation measure & error rate (classification) \\
\rule{0pt}{3ex} 

\multirow{5}{*}{Optimisation method} & SMAC, with executable = \\
&	smac-v2.06.01-development-619/smac \\
&	Initial Incumbent = Random \\
&	Execution Mode = SMAC \\
&	InitialN = 1 \\
\rule{0pt}{3ex} 

memLimit & 15 GB \\
\rule{0pt}{3ex} 

timeLimit & from 1,000s to 10,000s \\

\bottomrule
\end{tabular*}
\end{table}

None of the 4 meta-learning methods had their hyper-parameter values optimised to individual datasets. A more robust hyper-parameter optimisation procedure would be too time-consuming, given the very large number of experiments carried out in this work.


\subsection{Computational Environment and Runtime Limits} \label{sec:runtime-limits}

The experiments were executed in a Dual Intel 2.10GHz Xeon E5-2683 v4 Hexadeca-Core with 128GB RAM. In order to perform controlled experiments comparing different meta-learning methods with the same computational budget, recall that two types of experiments are performed, as reported in Section~\ref{sec:results}. 
The first type of experiment compares the results obtained by the three EAs (each evolving classification algorithms based on a single type of knowledge representation) with the results obtained by Auto-WEKA, which can recommend classification algorithms based on multiple knowledge representations. The second type of experiments compares the best EA (HEAD-DT, evolving decision-tree algorithms) against Auto-WEKA in an extended set of datasets.

In both types of experiments, to have a fair comparison among all meta-learning methods, each of them is allocated the same runtime limit. Experiments were performed with ten increasing values of the runtime limit for each meta-learning method, namely 1,000s (seconds), 2,000s, ..., up to 10,000s. These runtime limits refer to the time taken by a single run of each method on each dataset, on a single cross-validation fold. Due to space restrictions, the next section will report only the results for the smallest and the largest runtime limits, i.e., 1,000s and 10,000s. The results for the other runtime limits can be seen in~\cite{Basgalupp2018arxiv}.

In addition to the parameters that are common to all three EAs, which were set as described in Table~\ref{tab:parameters}, there is a parameter that is used by GGP-RI and HHEA-BNC, but not by HEAD-DT. This parameter is a timeout to evaluate each individual (candidate algorithm) of the EA. For GGP-RI, the value of this parameter starts with 10s (seconds) when the runtime limit for the entire run of GGP-RI is 1,000s. Then the individual evaluation timeout increases by 10s for each increase of 1,000s in GGP-RI's runtime, up to 100s, when the GGP-RI's runtime limit is 10,000s. For HHEA-BNC, the value of this parameter starts with 50s (seconds) when the runtime limit for the entire run of HHEA-BNC is 1,000s. Then the individual evaluation timeout increases by 50s for each increase of 1,000s in HHEA-BNC's runtime, up to 500s, when the HHEA-BNC's runtime limit is 10,000s. 
HEAD-DT does not need this parameter because the decision tree induction algorithms produced by this EA are relatively fast. The values of this parameter for HHEA-BNC are larger than the values for GGP-RI because the Bayesian network classification algorithms generated by the former tend to be considerably slower than the rule induction algorithms generated by the latter EA.

\section{Experimental Results} \label{sec:results}

This section presents the results of the following two types of experiments:

\begin{enumerate}

\item Experiments comparing four AutoML methods: the three EAs (HEAD-DT, GGP-RI, HHEA-BNC) and Auto-WEKA.

\item Experiments comparing one of the EAs (HEAD-DT, evolving decision tree algorithms) with Auto-WEKA, on an extended set of datasets.

\end{enumerate}

As mentioned earlier, due to the very large number of experiments, the first type of experiments use the 20 datasets shown in Table~\ref{tab:datasets}; whilst the second type of experiments uses an extended set of 40 datasets (the 20 datasets in Table~\ref{tab:datasets} plus the 20 datasets in Table~\ref{tab:additionalDatasets}).
We report results for the values of accuracy and Gmean (the geometric mean of sensitivity and specificity) for each dataset; and the average values of accuracy and GMean, as well as the average rank of each method based on these measures, over the corresponding datasets.
The lower the rank, the better the method. A method that outperforms every other method in every dataset has an average rank of 1.0 (first position). The complete tables with per-dataset results can be found in the Supplementary Results file.
Recall that, although we performed experiments with the runtime limit for meta-learning methods varying from 1,000 to 10,000 seconds, in increments of 1,000s, in general only the results for 1,000s and 10,000s are reported in this section, due to space restrictions. The results for the 10 different runtime limits can be found in the Supplementary Results file.


\subsection{Results Comparing Four AutoML Methods} \label{sec:general}

This section compares four types of AutoML methods, the three EAs and Auto-WEKA, in controlled experiments where all the four methods use the same runtime limit, as mentioned earlier.

Tables~\ref{tab:best-all-gmean1k} and \ref{tab:best-all-gmean10k} show the GMean results for each method, for the runtime limits of 1,000s and 10,000s, respectively. Recall that these runtime limits refer to a single run of a meta-learning method, for each fold of the cross-validation procedure. The last row of these tables show the average rank based on GMean over all 20 datasets. Tables~\ref{tab:best-all-acc1k} and \ref{tab:best-all-acc10k} show the accuracy results for each method, for the runtime limits of 1,000s and 10,000s, respectively.

\begin{table}[!htpb]
\centering \scriptsize
\caption{GMean results for the four AutoML methods (time limit: 1,000s).}
\label{tab:best-all-gmean1k}
\begin{tabular*}{\textwidth}
	{@{\extracolsep{\fill}}lcccc}
\toprule
Dataset & HEAD-DT & HHEA-BNC & GGP-RI & Auto-WEKA\\
\midrule
    CE    & 0.564 & 0.576 & 0.501 & \textbf{0.604} \\
    DM    & 0.559 & \textbf{0.596} & 0.523 & 0.557 \\
    MM    & 0.596 & \textbf{0.637} & 0.524 & 0.572 \\
    SC    & \textbf{0.535} & 0.497 & 0.392 & 0.471 \\
    DNA3  & 0.704 & \textbf{0.741} & 0.582 & 0.700 \\
    DNA11 & \textbf{0.568} & 0.544 & 0.498 & 0.506 \\
    PS    & \textbf{0.888} & 0.827 & 0.445 & 0.830 \\
    chen-2002 & 0.891 & 0.852 & 0.658 & \textbf{0.922} \\
    chowdary-2006 & 0.956 & 0.966 & 0.830 & \textbf{0.988} \\
    nutt-2003-v2 & 0.790 & 0.746 & 0.631 & \textbf{0.861} \\
    singh-2002 & 0.772 & 0.771 & 0.613 & \textbf{0.867} \\
    west-2001 & \textbf{0.913} & 0.886 & 0.617 & 0.888 \\
    dbworld-bodies & 0.725 & 0.753 & 0.582 & \textbf{0.765} \\
    dbworld-bodies-stemmed & 0.815 & 0.770 & 0.652 & \textbf{0.825} \\
    oh0.wc & 0.895 & \textbf{0.940} & 0.398 & 0.863 \\
    oh5.wc & 0.911 & \textbf{0.913} & 0.361 & 0.878 \\
    oh10.wc & 0.867 & \textbf{0.876} & 0.370 & 0.831 \\
    oh15.wc & 0.847 & \textbf{0.909} & 0.382 & 0.864 \\
    re0.wc & 0.831 & 0.841 & 0.489 & \textbf{0.849} \\
    re1.wc & \textbf{0.886} & 0.832 & 0.407 & 0.851 \\
\midrule
    Average & \textbf{0.776} & 0.774 & 0.523 & 0.775 \\
    Average Rank & \textbf{2.000} & \textbf{2.000} & 4.000 & \textbf{2.000} \\
        \bottomrule
    \end{tabular*}%
\end{table}%

\begin{table}[!htpb]
\centering \scriptsize
\caption{GMean results for the four AutoML methods (time limit: 10,000s).}
\label{tab:best-all-gmean10k}
\begin{tabular*}{\textwidth}
	{@{\extracolsep{\fill}}lcccc}
\toprule
Dataset & HEAD-DT & HHEA-BNC & GGP-RI & Auto-WEKA\\
\midrule
    CE    & 0.581 & 0.578 & 0.502 & \textbf{0.605} \\
    DM    & 0.517 & \textbf{0.629} & 0.544 & 0.544 \\
    MM    & 0.590 & \textbf{0.598} & 0.550 & 0.563 \\
    SC    & \textbf{0.559} & 0.528 & 0.389 & 0.454 \\
    DNA3  & 0.705 & \textbf{0.730} & 0.583 & 0.712 \\
    DNA11 & \textbf{0.578} & 0.497 & 0.506 & 0.524 \\
    PS    & \textbf{0.897} & 0.824 & 0.448 & 0.838 \\
    chen-2002 & 0.892 & 0.862 & 0.659 & \textbf{0.925} \\
    chowdary-2006 & 0.956 & 0.958 & 0.833 & \textbf{0.991} \\
    nutt-2003-v2 & 0.790 & 0.809 & 0.611 & \textbf{0.887} \\
    singh-2002 & 0.772 & 0.777 & 0.638 & \textbf{0.877} \\
    west-2001 & \textbf{0.913} & 0.879 & 0.624 & 0.878 \\
    dbworld-bodies & 0.725 & 0.784 & 0.585 & \textbf{0.816} \\
    dbworld-bodies-stemmed & 0.815 & 0.805 & 0.649 & \textbf{0.892} \\
    oh0.wc & 0.893 & \textbf{0.918} & 0.398 & 0.884 \\
    oh5.wc & \textbf{0.914} & 0.896 & 0.364 & 0.880 \\
    oh10.wc & \textbf{0.864} & 0.847 & 0.369 & 0.835 \\
    oh15.wc & 0.859 & \textbf{0.900} & 0.381 & 0.867 \\
    re0.wc & 0.831 & 0.827 & 0.489 & \textbf{0.841} \\
    re1.wc & \textbf{0.894} & 0.883 & 0.407 & 0.859 \\
\midrule
    Average & 0.777 & 0.777 & 0.526 & \textbf{0.783} \\
    Average Rank & 2.050 & 2.100 & 3.850 & \textbf{2.000} \\
    \bottomrule
    \end{tabular*}%
\end{table}%

In Table~\ref{tab:best-all-gmean1k}, with GMean results for the smallest runtime limit of 1,000s, the best average ranks were jointly obtained by three methods, HEAD-DT, HHEA-BNC and Auto-WEKA; whilst HEAD-DT obtained a slightly better average GMean value.
In Table~\ref{tab:best-all-gmean10k}, with results for the longest runtime limit of 10,000s, Auto-WEKA obtained a slightly better result (regarding both the average rank and the average GMean value) than HEAD-DT and HHEA-BNC. 
In both tables, GGP-RI was clearly the worst performing method. This result seem partly due to the fact that GGP-RI had poor results in many datasets with a large number of numerical attributes.
Comparing the average GMean values of each method across both tables, one can observe that the three EAs have only slightly improved their GMean values from 1,000s to 10,000s -- an improvement of just 0.001 for HEAD-DT and 0.003 for the other two EAs. By contrast, Auto-WEKA obtained a somewhat greater GMean improvement of 0.008, when the runtime limit increased from 1,000s to 10,000s. 

Hence, Auto-WEKA has benefited from the increase in runtime limit more than the EAs. This seems due to the fact that Auto-WEKA is searching in a much more diverse space of classification algorithms, in terms of knowledge representations. Recall that each EA's search space includes algorithms from a single knowledge representation (decision trees, if-then classification rules or Bayesian network classifiers), whilst Auto-WEKA's search space includes 33 classification algorithms from multiple types of knowledge representation. Hence, it seems natural that Auto-WEKA requires more time to find the best type of algorithm to be recommended.

When analyzing the results for the accuracy measure, the scenario changes a little. It is possible to see, in both Table~\ref{tab:best-all-acc1k} and Table~\ref{tab:best-all-acc10k}, for the runtime limits of 1,000s and 10,000s, respectively, there is a clearer difference in relative ranks of the three best methods. More precisely, when predictive accuracy is evaluated by the accuracy measure, Auto-WEKA is the best method, followed by HHEA-BNC and HEAD-DT in second in third places, respectively, in terms of average rank. In terms of average accuracy, HHEA-BNC and Auto-WEKA obtain the joint best result in Table~\ref{tab:best-all-acc1k} (1,000s), but Auto-WEKA is again the clear winner in Table~\ref{tab:best-all-acc10k} (10,000s). Again, Auto-WEKA was the method that most benefited from the increase in the runtime limit, with a small improvement of average accuracy, namely 0.008.
Again, GGP-RI was clearly the worst performing method.

To explain these results, recall that Auto-WEKA explicitly optimizes the accuracy measure when searching for the best algorithm configuration, whereas the EAs are optimizing the F-Measure. Hence, it is natural that Auto-WEKA obtains the best predictive performance when the results are evaluated by the Accuracy measure.

\begin{table}[!htpb]
\centering \scriptsize
\caption{Accuracy results for the four AutoML methods (time limit: 1,000s).}
\label{tab:best-all-acc1k}
\begin{tabular*}{\textwidth}
	{@{\extracolsep{\fill}}lcccc}
\toprule
Dataset & HEAD-DT & HHEA-BNC & GGP-RI & Auto-WEKA \\
\midrule
    CE    & 0.613 & 0.615 & 0.478 & \textbf{0.649} \\
    DM    & 0.637 & \textbf{0.708} & 0.598 & 0.672 \\
    MM    & 0.720 & \textbf{0.748} & 0.641 & 0.722 \\
    SC    & 0.826 & 0.802 & 0.764 & \textbf{0.828} \\
    DNA3  & 0.846 & 0.841 & 0.760 & \textbf{0.856} \\
    DNA11 & \textbf{0.752} & 0.743 & 0.682 & 0.708 \\
    PS    & \textbf{0.982} & 0.978 & 0.933 & 0.975 \\
    chen-2002 & 0.896 & 0.867 & 0.663 & \textbf{0.926} \\
    chowdary-2006 & 0.959 & 0.971 & 0.832 & \textbf{0.991} \\
    nutt-2003-v2 & 0.760 & 0.730 & 0.537 & \textbf{0.840} \\
    singh-2002 & 0.772 & 0.771 & 0.539 & \textbf{0.867} \\
    west-2001 & \textbf{0.910} & 0.888 & 0.511 & 0.880 \\
    dbworld-bodies & 0.721 & \textbf{0.764} & 0.523 & \textbf{0.764} \\
    dbworld-bodies-stemmed & 0.806 & 0.783 & 0.610 & \textbf{0.825} \\
    oh0.wc & 0.825 & \textbf{0.896} & 0.070 & 0.778 \\
    oh5.wc & 0.846 & \textbf{0.848} & 0.048 & 0.789 \\
    oh10.wc & 0.777 & \textbf{0.790} & 0.053 & 0.721 \\
    oh15.wc & 0.746 & \textbf{0.844} & 0.061 & 0.774 \\
    re0.wc & 0.755 & 0.760 & 0.240 & \textbf{0.783} \\
    re1.wc & \textbf{0.807} & 0.742 & 0.084 & 0.755 \\
\midrule
    Average & 0.798 & \textbf{0.805} & 0.481 & \textbf{0.805} \\
    Average Rank & 2.150 & 2.025 & 4.000 & \textbf{1.825} \\
        \bottomrule
    \end{tabular*}%
\end{table}%

\begin{table}[!htpb]
\centering \scriptsize
\caption{Accuracy results for the four AutoML methods (time limit: 10,000s).}
\label{tab:best-all-acc10k}
\begin{tabular*}{\textwidth}
	{@{\extracolsep{\fill}}lcccc}
\toprule
Dataset & HEAD-DT & HHEA-BNC & GGP-RI & Auto-WEKA\\
\midrule
    CE    & 0.623 & 0.614 & 0.482 & \textbf{0.649} \\
    DM    & 0.604 & \textbf{0.713} & 0.615 & 0.665 \\
    MM    & 0.702 & \textbf{0.730} & 0.658 & 0.706 \\
    SC    & 0.818 & 0.806 & 0.764 & \textbf{0.826} \\
    DNA3  & 0.847 & 0.838 & 0.758 & \textbf{0.855} \\
    DNA11 & \textbf{0.747} & 0.730 & 0.685 & 0.701 \\
    PS    & \textbf{0.984} & 0.977 & 0.933 & 0.975 \\
    chen-2002 & 0.896 & 0.868 & 0.666 & \textbf{0.927} \\
    chowdary-2006 & 0.959 & 0.965 & 0.837 & \textbf{0.993} \\
    nutt-2003-v2 & 0.760 & 0.790 & 0.517 & \textbf{0.873} \\
    singh-2002 & 0.772 & 0.777 & 0.574 & \textbf{0.877} \\
    west-2001 & \textbf{0.910} & 0.883 & 0.538 & 0.868 \\
    dbworld-bodies & 0.721 & 0.792 & 0.530 & \textbf{0.812} \\
    dbworld-bodies-stemmed & 0.806 & 0.814 & 0.605 & \textbf{0.891} \\
    oh0.wc & 0.824 & \textbf{0.868} & 0.071 & 0.809 \\
    oh5.wc & \textbf{0.850} & 0.829 & 0.051 & 0.793 \\
    oh10.wc & \textbf{0.773} & 0.754 & 0.056 & 0.727 \\
    oh15.wc & 0.764 & \textbf{0.832} & 0.060 & 0.778 \\
    re0.wc & 0.752 & 0.746 & 0.240 & \textbf{0.774} \\
    re1.wc & \textbf{0.820} & 0.802 & 0.084 & 0.767 \\
\midrule
    Average & 0.797 & 0.806 & 0.486 & \textbf{0.813} \\
    Average Rank & 2.150 & 2.050 & 3.950 & \textbf{1.850} \\
    \bottomrule
    \end{tabular*}%
\end{table}%

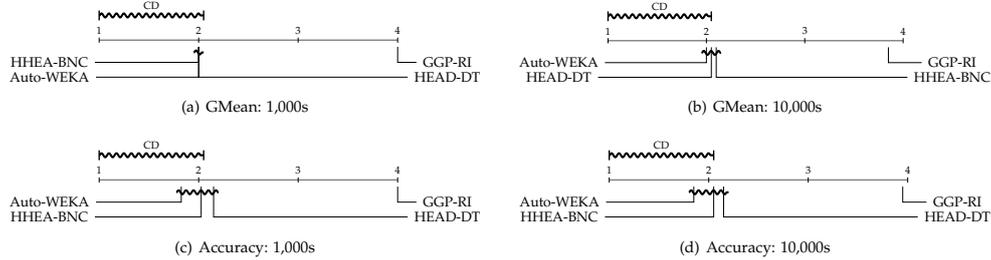
\begin{figure}[!htpb] 
\centering 
\resizebox{.45\columnwidth}{!}{
\subfigure[GMean: 1,000s]{
\begin{tikzpicture}[xscale=2]
\node (Label) at  (01.5250,0.7) {\tiny{CD}}; 
\draw[decorate,decoration={snake,amplitude=.4mm,segment length=1.5mm,post length=0mm}, very thick, color = black](01.0000, 0.5) -- (02.0500, 0.5);
\foreach \x in {01.0000,02.0500} \draw[thick,color = black] (\x, 0.4) -- (\x, 0.6);

\draw[gray, thick](01.0000, 0) -- (04.0000, 0);
\foreach \x in {01.0000,02.0000,03.0000,04.0000}\draw (\x cm,1.5pt) -- (\x cm, -1.5pt);
\node (Label) at (01.0000,0.2) {\tiny{1}};
\node (Label) at (02.0000,0.2) {\tiny{2}};
\node (Label) at (03.0000,0.2) {\tiny{3}};
\node (Label) at (04.0000,0.2) {\tiny{4}};
\draw[decorate,decoration={snake,amplitude=.4mm,segment length=1.5mm,post length=0mm}, very thick, color = black](01.9500,-00.2500) -- ( 02.0500,-00.2500);
\node (Point) at (02.0000, 0){};  \node (Label) at (0.5,-00.4500){\scriptsize{HHEA-BNC}}; \draw (Point) |- (Label);
\node (Point) at (02.0000, 0){};  \node (Label) at (0.5,-00.7500){\scriptsize{Auto-WEKA}}; \draw (Point) |- (Label);
\node (Point) at (04.0000, 0){};  \node (Label) at (4.5,-00.4500){\scriptsize{GGP-RI}}; \draw (Point) |- (Label);
\node (Point) at (02.0000, 0){};  \node (Label) at (4.5,-00.7500){\scriptsize{HEAD-DT}}; \draw (Point) |- (Label);
\end{tikzpicture}
}
}
\resizebox{.45\columnwidth}{!}{
\subfigure[GMean: 10,000s]{
\begin{tikzpicture}[xscale=2]
\node (Label) at  (01.5250,0.7) {\tiny{CD}}; 
\draw[decorate,decoration={snake,amplitude=.4mm,segment length=1.5mm,post length=0mm}, very thick, color = black](01.0000, 0.5) -- (02.0500, 0.5);
\foreach \x in {01.0000,02.0500} \draw[thick,color = black] (\x, 0.4) -- (\x, 0.6);

\draw[gray, thick](01.0000, 0) -- (04.0000, 0);
\foreach \x in {01.0000,02.0000,03.0000,04.0000}\draw (\x cm,1.5pt) -- (\x cm, -1.5pt);
\node (Label) at (01.0000,0.2) {\tiny{1}};
\node (Label) at (02.0000,0.2) {\tiny{2}};
\node (Label) at (03.0000,0.2) {\tiny{3}};
\node (Label) at (04.0000,0.2) {\tiny{4}};
\draw[decorate,decoration={snake,amplitude=.4mm,segment length=1.5mm,post length=0mm}, very thick, color = black](01.9500,-00.2500) -- ( 02.1500,-00.2500);
\node (Point) at (02.0000, 0){};  \node (Label) at (0.5,-00.4500){\scriptsize{Auto-WEKA}}; \draw (Point) |- (Label);
\node (Point) at (02.0500, 0){};  \node (Label) at (0.5,-00.7500){\scriptsize{HEAD-DT}}; \draw (Point) |- (Label);
\node (Point) at (03.8500, 0){};  \node (Label) at (4.5,-00.4500){\scriptsize{GGP-RI}}; \draw (Point) |- (Label);
\node (Point) at (02.1000, 0){};  \node (Label) at (4.5,-00.7500){\scriptsize{HHEA-BNC}}; \draw (Point) |- (Label);
\end{tikzpicture}
}
}

\resizebox{.45\columnwidth}{!}{
\subfigure[Accuracy: 1,000s]{
\begin{tikzpicture}[xscale=2]
\node (Label) at  (01.5250,0.7) {\tiny{CD}}; 
\draw[decorate,decoration={snake,amplitude=.4mm,segment length=1.5mm,post length=0mm}, very thick, color = black](01.0000, 0.5) -- (02.0500, 0.5);
\foreach \x in {01.0000,02.0500} \draw[thick,color = black] (\x, 0.4) -- (\x, 0.6);

\draw[gray, thick](01.0000, 0) -- (04.0000, 0);
\foreach \x in {01.0000,02.0000,03.0000,04.0000}\draw (\x cm,1.5pt) -- (\x cm, -1.5pt);
\node (Label) at (01.0000,0.2) {\tiny{1}};
\node (Label) at (02.0000,0.2) {\tiny{2}};
\node (Label) at (03.0000,0.2) {\tiny{3}};
\node (Label) at (04.0000,0.2) {\tiny{4}};
\draw[decorate,decoration={snake,amplitude=.4mm,segment length=1.5mm,post length=0mm}, very thick, color = black](01.7750,-00.2500) -- ( 02.2000,-00.2500);
\node (Point) at (01.8250, 0){};  \node (Label) at (0.5,-00.4500){\scriptsize{Auto-WEKA}}; \draw (Point) |- (Label);
\node (Point) at (02.0250, 0){};  \node (Label) at (0.5,-00.7500){\scriptsize{HHEA-BNC}}; \draw (Point) |- (Label);
\node (Point) at (04.0000, 0){};  \node (Label) at (4.5,-00.4500){\scriptsize{GGP-RI}}; \draw (Point) |- (Label);
\node (Point) at (02.1500, 0){};  \node (Label) at (4.5,-00.7500){\scriptsize{HEAD-DT}}; \draw (Point) |- (Label);
\end{tikzpicture}
}
}
\resizebox{.45\columnwidth}{!}{
\subfigure[Accuracy: 10,000s]{
\begin{tikzpicture}[xscale=2]
\node (Label) at  (01.5250,0.7) {\tiny{CD}}; 
\draw[decorate,decoration={snake,amplitude=.4mm,segment length=1.5mm,post length=0mm}, very thick, color = black](01.0000, 0.5) -- (02.0500, 0.5);
\foreach \x in {01.0000,02.0500} \draw[thick,color = black] (\x, 0.4) -- (\x, 0.6);

\draw[gray, thick](01.0000, 0) -- (04.0000, 0);
\foreach \x in {01.0000,02.0000,03.0000,04.0000}\draw (\x cm,1.5pt) -- (\x cm, -1.5pt);
\node (Label) at (01.0000,0.2) {\tiny{1}};
\node (Label) at (02.0000,0.2) {\tiny{2}};
\node (Label) at (03.0000,0.2) {\tiny{3}};
\node (Label) at (04.0000,0.2) {\tiny{4}};
\draw[decorate,decoration={snake,amplitude=.4mm,segment length=1.5mm,post length=0mm}, very thick, color = black](01.8000,-00.2500) -- ( 02.2000,-00.2500);
\node (Point) at (01.8500, 0){};  \node (Label) at (0.5,-00.4500){\scriptsize{Auto-WEKA}}; \draw (Point) |- (Label);
\node (Point) at (02.0500, 0){};  \node (Label) at (0.5,-00.7500){\scriptsize{HHEA-BNC}}; \draw (Point) |- (Label);
\node (Point) at (03.9500, 0){};  \node (Label) at (4.5,-00.4500){\scriptsize{GGP-RI}}; \draw (Point) |- (Label);
\node (Point) at (02.1500, 0){};  \node (Label) at (4.5,-00.7500){\scriptsize{HEAD-DT}}; \draw (Point) |- (Label);
\end{tikzpicture}
}
}
\caption{Critical diagrams showing average GMean/Accuracy ranks and Nemenyi's critical difference (CD) for the four AutoML methods.}
\label{fig:CD-gmean}
\end{figure}

Figure~\ref{fig:CD-gmean} shows the critical diagrams comparing the four AutoML methods in terms of their average rank based on both GMean (in the top two diagrams) and accuracy (in the bottom two diagrams). For both measures, and for both the runtime limits of 1,000s and 10,000s, we can see that there is no statistically-significant difference among all methods, with the exception of GGP-RI, which is significantly outperformed by the other three methods.

\begin{figure}[!htpb] \centering 
\resizebox{0.49\columnwidth}{!}{
\subfigure[\normalsize GMean]{
\includegraphics[width=\linewidth]{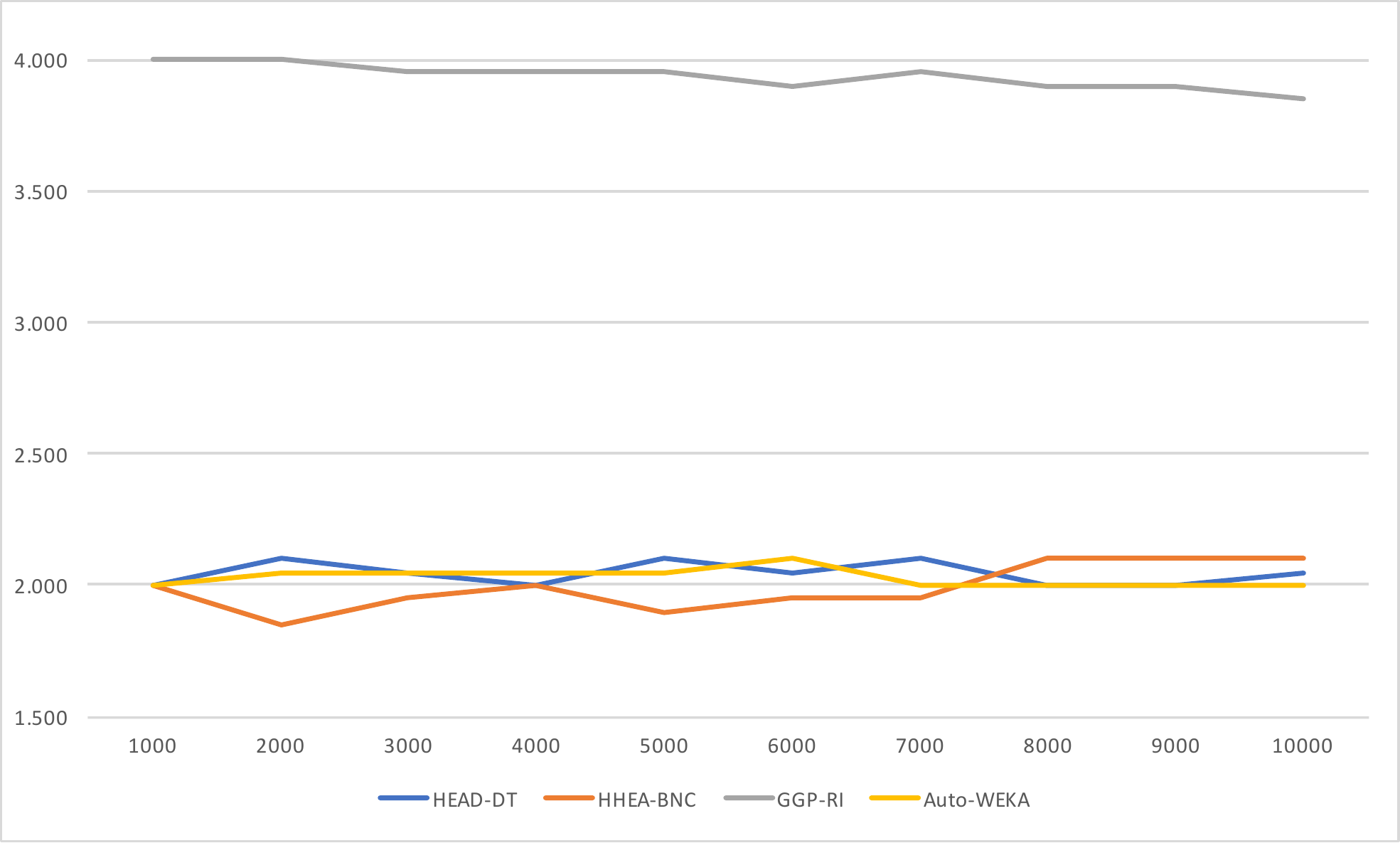} \label{fig:all-ranks-gmean}
}}
\resizebox{0.49\columnwidth}{!}{
\subfigure[\normalsize Accuracy]{
\includegraphics[width=\linewidth]{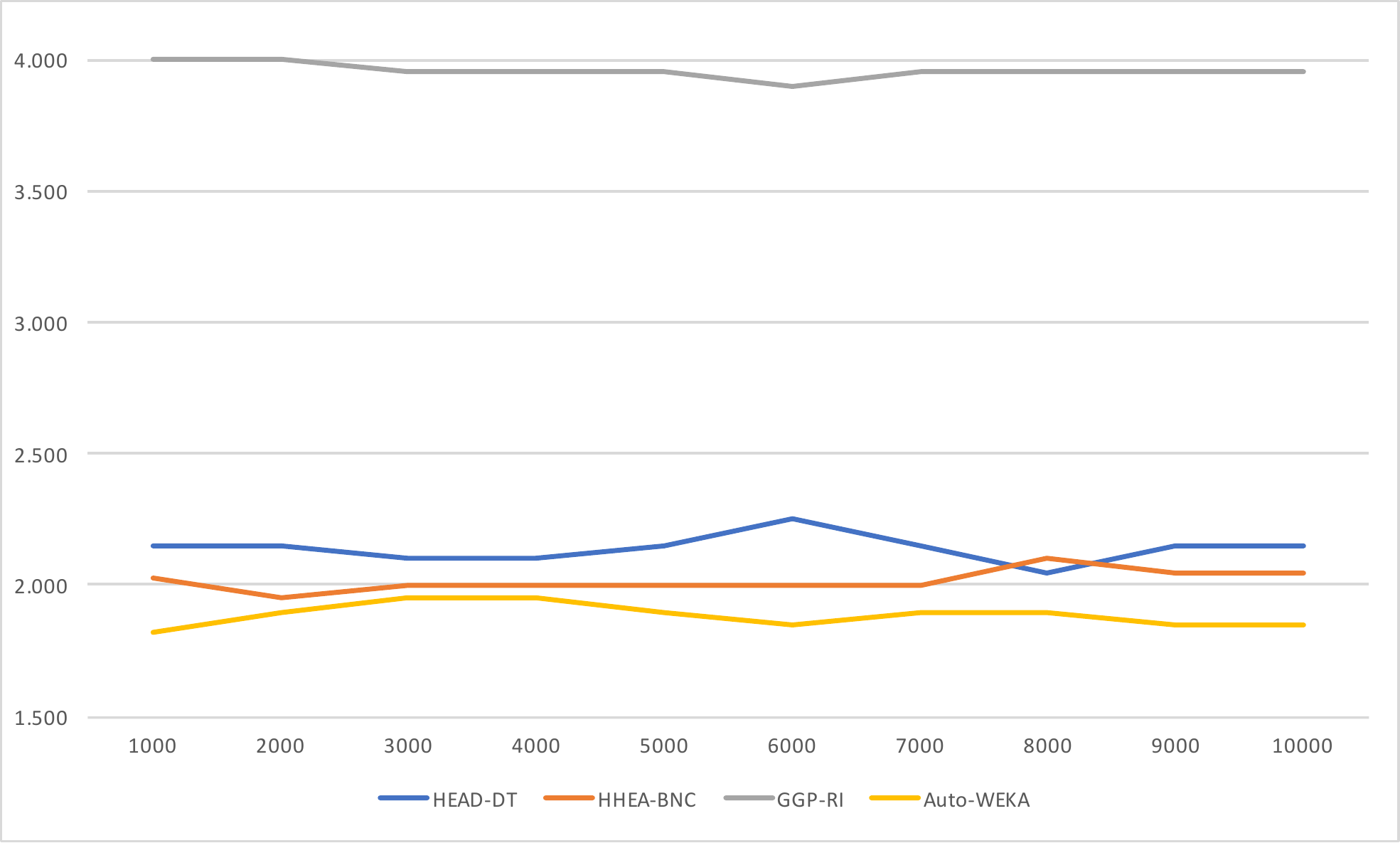} \label{fig:all-ranks-acc}
}}
\caption{Evolution of average ranks for all AutoML methods across the 10 runtime limits.}
\label{fig:all-ranks}
\end{figure}

As mentioned earlier, the analysis of the results so far focused only on the runtime limits of 1,000s and 10,000s due to space restrictions, but we performed experiments with 10 different limits (from 1,000s up to 10,000s). Figure~\ref{fig:all-ranks-gmean} shows the evolution of the GMean average ranks for the four meta-learning methods across the 10 runtime limits. This figure shows that HHEA-BNC tends to achieve overall the best (lowest) average rank until the runtime limit of 7,000s, whilst for longer runtime limits Auto-WEKA and HEAD-DT tend to share the best rank, with Auto-WEKA slightly better at the last runtime limit.

Figure~\ref{fig:all-ranks-acc} shows the same evolution, but this time regarding average accuracy ranks. 
In this case, Auto-WEKA remains the best method across all runtime limits, and for nearly all runtime limits, the second place is obtained by HHEA-BNC.
Note that GGP-RI remained consistently the worst method across all 10 runtime limits, for both GMean and accuracy results.

\begin{figure}[!htpb] \centering 
\resizebox{0.7\columnwidth}{!}{
\subfigure[\normalsize1,000s]{
\includegraphics[width=\linewidth]{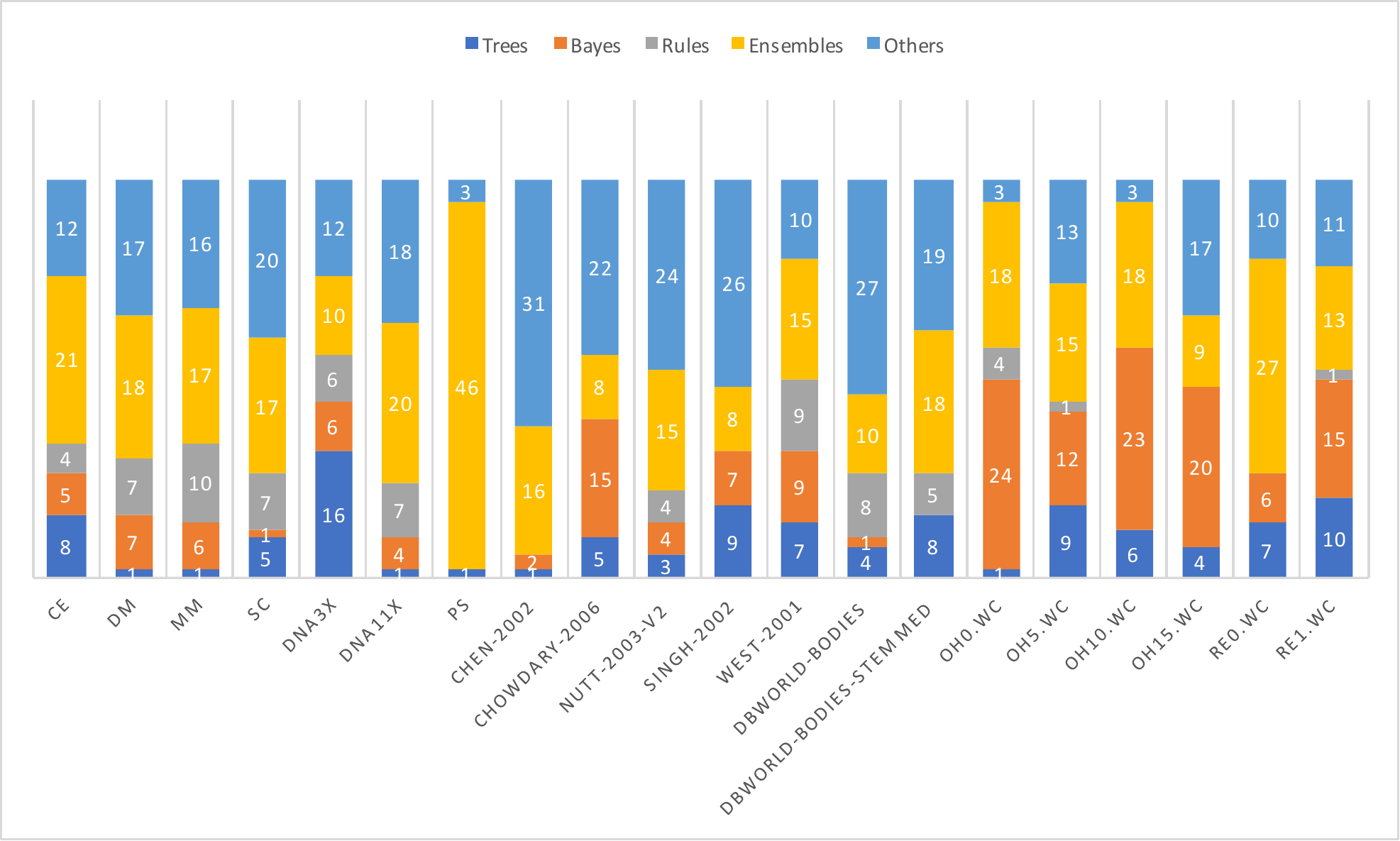}
}}

\resizebox{0.7\columnwidth}{!}{
\subfigure[\normalsize10,000s]{
\includegraphics[width=\linewidth]{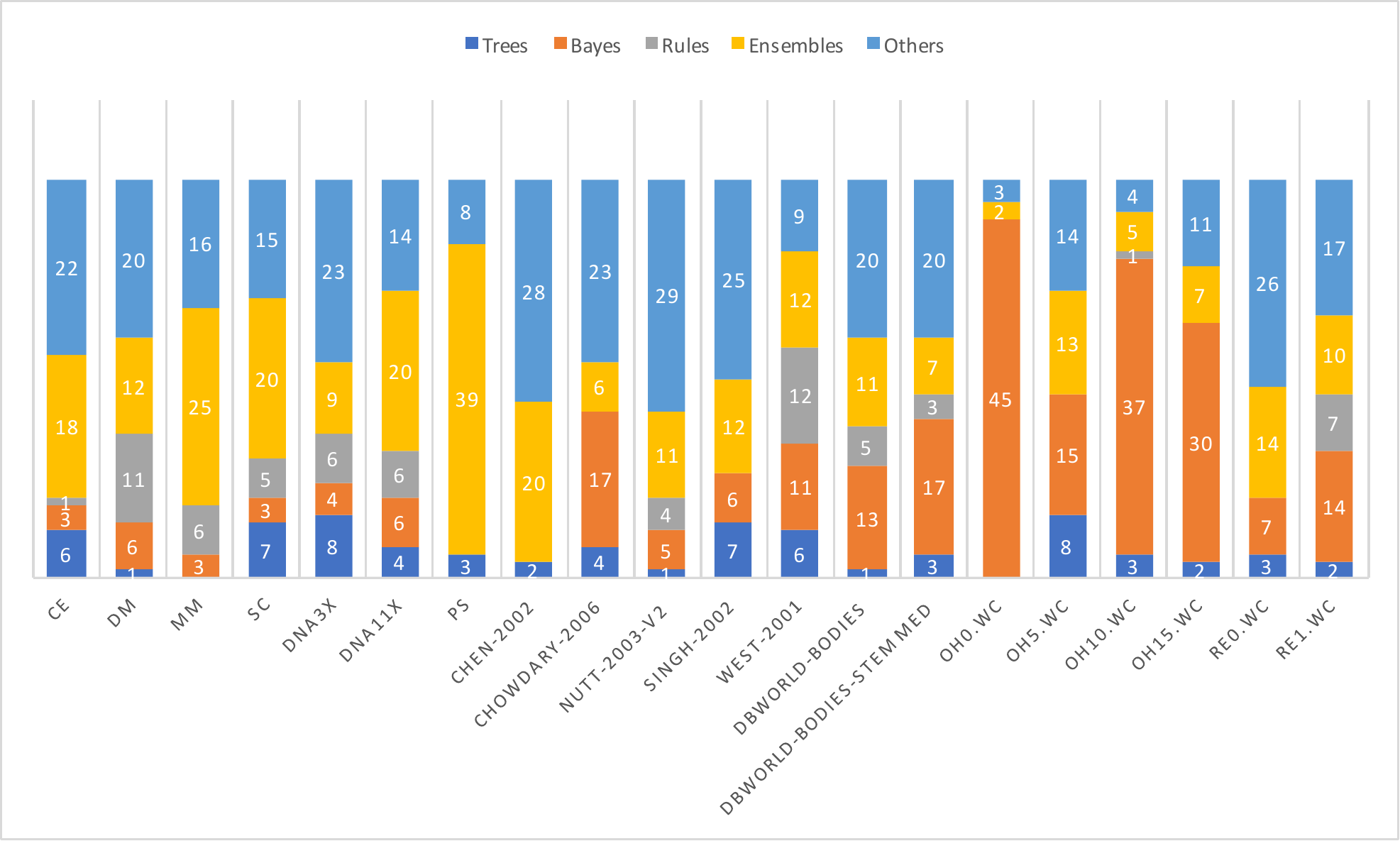}
}}
\caption{Number of times each type of classification algorithm is selected by Auto-WEKA.}
\label{fig:AW-models}
\end{figure}

Figures~\ref{fig:AW-models}(a) and \ref{fig:AW-models}(b) show the broad types of algorithms recommended by Auto-WEKA per dataset, for the runtime limits of 1,000s and 10,000s, respectively. Since Auto-WEKA considers a large number of algorithms, instead of referring to specific algorithms, the graphs show the frequency of recommendations for five broad types of algorithms, namely: the three types of algorithms that are considered by the three EAs (decision trees, if-then classification rules, and Bayesian network classifiers), ensemble methods and all the others. Note that the variability of the selected types of algorithms is high, highlighting the difficulty of  selecting the best algorithm for each dataset.

For the runtime limit of 1,000s (Figure~\ref{fig:AW-models}(a)), ensembles had the highest prevalence across the datasets; they were selected by Auto-WEKA in 33.9\% of the cases, closely followed by decision-tree algorithms, selected in 31.4\% of the cases. 
For the runtime limit of 10,000s (Figure~\ref{fig:AW-models}(b)), these two types of classification algorithms swapped places in the ranking by prevalence, i.e., decision-tree algorithms were selected by Auto-WEKA in 34.7\% of the cases, whilst ensembles were selected in 27.3\% of the cases. 
Bayesian classification algorithms also did relatively well, partly because they had a high prevalence among the text mining datasets. For both runtime limits, Bayesian classification algorithms were the third most selected type of classification algorithm: they were selected in 16.7\% of the cases in Figure~\ref{fig:AW-models}(a) and in 24.2\% of the cases in Figure~\ref{fig:AW-models}(b).
For both runtime limits, rule induction algorithms had small frequencies of selection, only 7.9\% in Figure~\ref{fig:AW-models}(a) and 6.7\% in Figure~\ref{fig:AW-models}(b). This is consistent with the fact that, out of the 3 EAs for AutoML evaluated in this work, GGP-RI (which evolved rule induction algorithms) obtained clearly the worst result.

\subsection{More extensive experiments comparing HEAD-DT and Auto-WEKA}

In this section we compare HEAD-DT and Auto-WEKA in an extended set of 40 datasets. This includes the 20 datasets used in the previous section plus 20 other datasets, as discussed in Section~\ref{sec:datasets}. As mentioned earlier, the motivation for using this larger set of datasets only to compare the two methods in this section, rather than to compare more methods in the previous section, is the much larger amount of time associated with the experiments using all the 40 datasets. 
This section uses the same experimental methodology used in the previous section, using 10-fold cross-validation and comparing the two methods with the same runtime limit, varying this limit from 1,000s to 10,000s, in increments of 1,000s. Again, due to space restrictions, we report results only for the smallest and longest runtime limits, namely 1,000s and 10,000s; but the results for the 10 different runtime limits can be found in the Supplementary Results file.

\begin{table}[htbp]
  \centering \scriptsize
\caption{Accuracy results for HEAD-DT and Auto-WEKA (time limits: 1,000s and 10,000s).}
\label{tab:newAcc}
  \begin{tabular*}{\textwidth}
	{@{\extracolsep{\fill}}lcccc}
\toprule
          & \multicolumn{2}{c}{1,000s} & \multicolumn{2}{c}{10,000s} \\
          & \multicolumn{1}{l}{HEAD-DT} & \multicolumn{1}{l}{Auto-WEKA} & \multicolumn{1}{l}{HEAD-DT} & \multicolumn{1}{l}{Auto-WEKA} \\
\midrule
    CE    & 0.613 & \textbf{0.649} & 0.623 & \textbf{0.649} \\
    DM    & 0.637 & \textbf{0.672} & 0.604 & \textbf{0.665} \\
    MM    & 0.720 & \textbf{0.722} & 0.702 & \textbf{0.706} \\
    SC    & 0.826 & \textbf{0.828} & 0.818 & \textbf{0.826} \\
    DNA3 & 0.846 & \textbf{0.856} & 0.847 & \textbf{0.855} \\
    DNA11 & \textbf{0.752} & 0.708 & \textbf{0.747} & 0.701 \\
    PS    & \textbf{0.982} & 0.975 & \textbf{0.984} & 0.975 \\
    chen-2002 & 0.896 & \textbf{0.926} & 0.896 & \textbf{0.927} \\
    chowdary-2006 & 0.959 & \textbf{0.991} & 0.959 & \textbf{0.993} \\
    nutt-2003-v2 & 0.760 & \textbf{0.840} & 0.760 & \textbf{0.873} \\
    singh-2002 & 0.772 & \textbf{0.867} & 0.772 & \textbf{0.877} \\
    west-2001 & \textbf{0.910} & 0.880 & \textbf{0.910} & 0.868 \\
    dbworld-bodies & 0.721 & \textbf{0.764} & 0.721 & \textbf{0.812} \\
    dbworld-bodies-stemmed & 0.806 & \textbf{0.825} & 0.806 & \textbf{0.891} \\
    oh0.wc & \textbf{0.825} & 0.778 & \textbf{0.824} & 0.809 \\
    oh5.wc & \textbf{0.846} & 0.789 & \textbf{0.850} & 0.793 \\
    oh10.wc & \textbf{0.777} & 0.721 & \textbf{0.773} & 0.727 \\
    oh15.wc & 0.746 & \textbf{0.774} & 0.764 & \textbf{0.778} \\
    re0.wc & 0.755 & \textbf{0.783} & 0.752 & \textbf{0.774} \\
    re1.wc & \textbf{0.807} & 0.755 & \textbf{0.820} & 0.767 \\
    abalone & \textbf{0.265} & 0.263 & \textbf{0.269} & 0.263 \\
    car   & 0.984 & \textbf{0.994} & 0.983 & \textbf{0.997} \\
    convex & \textbf{0.712} & 0.531 & \textbf{0.714} & 0.531 \\
    germancredit & \textbf{0.750} & 0.738 & \textbf{0.750} & 0.739 \\
    krvskp & \textbf{0.995} & 0.962 & \textbf{0.995} & 0.962 \\
    madelon & \textbf{0.781} & 0.735 & 0.768 & \textbf{0.784} \\
    mnist & 0.886 & \textbf{0.929} & 0.887 & \textbf{0.934} \\
    mnistrotationbackimagenew & \textbf{0.343} & 0.214 & \textbf{0.343} & 0.225 \\
    secom & 0.932 & 0.932 & 0.931 & \textbf{0.933} \\
    semeion & 0.763 & \textbf{0.894} & 0.758 & \textbf{0.907} \\
    shuttle & \textbf{1.000} & 0.999 & \textbf{1.000} & 0.999 \\
    waveform & 0.760 & \textbf{0.868} & 0.763 & \textbf{0.868} \\
    winequalitywhite & 0.622 & \textbf{0.676} & 0.627 & \textbf{0.672} \\
    yeast & 0.584 & \textbf{0.602} & 0.582 & \textbf{0.607} \\
    sick & \textbf{0.989} & 0.978 & \textbf{0.989} & 0.980 \\
    splice & \textbf{0.990} & 0.949 & \textbf{0.988} & 0.955 \\
    kropt & \textbf{0.796} & 0.680 & \textbf{0.801} & 0.761 \\
    quake & 0.535 & \textbf{0.553} & 0.529 & \textbf{0.546} \\
    pc4 & 0.889 & \textbf{0.891} & 0.886 & \textbf{0.896} \\
    magicTelescope & \textbf{0.852} & 0.831 & \textbf{0.853} & 0.840 \\
    \midrule
    Average & \textbf{0.785} & 0.783 & 0.784 & \textbf{0.792} \\
    \# wins & 18    & \textbf{21} & 17    & \textbf{23} \\
    \bottomrule
    \end{tabular*}%
\end{table}%

\begin{table}[htbp]
  \centering \scriptsize
\caption{GMean results for HEAD-DT and Auto-WEKA (time limits: 1,000s and 10,000s).}
\label{tab:newGmean}
  \begin{tabular*}{\textwidth}
	{@{\extracolsep{\fill}}lcccc}
\toprule
          & \multicolumn{2}{c}{1,000s} & \multicolumn{2}{c}{10,000s} \\
          & \multicolumn{1}{l}{HEAD-DT} & \multicolumn{1}{l}{Auto-WEKA} & \multicolumn{1}{l}{HEAD-DT} & \multicolumn{1}{l}{Auto-WEKA} \\
\midrule
    CE    & 0.564 & \textbf{0.604} & 0.581 & \textbf{0.605} \\
    DM    & \textbf{0.559} & 0.557 & 0.517 & \textbf{0.544} \\
    MM    & \textbf{0.596} & 0.572 & \textbf{0.590} & 0.563 \\
    SC    & \textbf{0.535} & 0.471 & \textbf{0.559} & 0.454 \\
    DNA3 & \textbf{0.704} & 0.700 & 0.705 & \textbf{0.712} \\
    DNA11 & \textbf{0.568} & 0.506 & \textbf{0.578} & 0.524 \\
    PS    & \textbf{0.888} & 0.830 & \textbf{0.897} & 0.838 \\
    chen-2002 & 0.891 & \textbf{0.922} & 0.892 & \textbf{0.925} \\
    chowdary-2006 & 0.956 & \textbf{0.988} & 0.956 & \textbf{0.991} \\
    nutt-2003-v2 & 0.790 & \textbf{0.861} & 0.790 & \textbf{0.887} \\
    singh-2002 & 0.772 & \textbf{0.867} & 0.772 & \textbf{0.877} \\
    west-2001 & \textbf{0.913} & 0.888 & \textbf{0.913} & 0.878 \\
    dbworld-bodies & 0.725 & \textbf{0.765} & 0.725 & \textbf{0.816} \\
    dbworld-bodies-stemmed & 0.815 & \textbf{0.825} & 0.815 & \textbf{0.892} \\
    oh0.wc & \textbf{0.895} & 0.863 & \textbf{0.893} & 0.884 \\
    oh5.wc & \textbf{0.911} & 0.878 & \textbf{0.914} & 0.880 \\
    oh10.wc & \textbf{0.867} & 0.831 & \textbf{0.864} & 0.835 \\
    oh15.wc & 0.847 & \textbf{0.864} & 0.859 & \textbf{0.867} \\
    re0.wc & 0.831 & \textbf{0.849} & 0.831 & \textbf{0.841} \\
    re1.wc & \textbf{0.886} & 0.851 & \textbf{0.894} & 0.859 \\
    abalone & \textbf{0.486} & 0.483 & \textbf{0.489} & 0.483 \\
    car   & 0.987 & \textbf{0.996} & 0.987 & \textbf{0.998} \\
    convex & \textbf{0.712} & 0.531 & \textbf{0.714} & 0.531 \\
    germancredit & \textbf{0.655} & 0.630 & \textbf{0.657} & 0.638 \\
    krvskp & \textbf{0.995} & 0.961 & \textbf{0.995} & 0.962 \\
    madelon & \textbf{0.781} & 0.735 & 0.768 & \textbf{0.784} \\
    mnist & 0.935 & \textbf{0.960} & 0.936 & \textbf{0.963} \\
    mnistrotationbackimagenew & \textbf{0.564} & 0.441 & \textbf{0.564} & 0.452 \\
    secom & 0.254 & \textbf{0.274} & 0.256 & \textbf{0.268} \\
    semeion & 0.862 & \textbf{0.940} & 0.859 & \textbf{0.947} \\
    shuttle & \textbf{1.000} & 0.997 & \textbf{1.000} & 0.997 \\
    waveform & 0.818 & \textbf{0.900} & 0.820 & \textbf{0.900} \\
    winequalitywhite & 0.712 & \textbf{0.738} & 0.716 & \textbf{0.728} \\
    yeast & 0.708 & \textbf{0.721} & 0.706 & \textbf{0.724} \\
    sick & \textbf{0.940} & 0.888 & \textbf{0.943} & 0.894 \\
    splice & \textbf{0.993} & 0.961 & \textbf{0.991} & 0.965 \\
    kropt & \textbf{0.881} & 0.808 & \textbf{0.884} & 0.858 \\
    quake & \textbf{0.516} & 0.498 & \textbf{0.506} & 0.496 \\
    pc4 & \textbf{0.689} & 0.550 & \textbf{0.694} & 0.579 \\
    magicTelescope & \textbf{0.821} & 0.793 & \textbf{0.823} & 0.802 \\
\midrule
    Average & \textbf{0.770} & 0.757 & \textbf{0.771} & 0.766 \\
    \# wins & \textbf{24} & 16    & \textbf{21}    & 19 \\
\bottomrule
    \end{tabular*}%
\end{table}%

Table~\ref{tab:newAcc} and Table~\ref{tab:newGmean} show the accuracy and GMean values, respectively, obtained by HEAD-DT and Auto-WEKA with the runtime limits of 1,000s and 10,000s. In terms of accuracy, Auto-WEKA has somewhat outperformed HEAD-DT overall, whilst the opposite was observed for the GMean measure. This result is consistent with the fact that Auto-WEKA's search tries to optimize the accuracy measure (unlike HEAD-DT), as discussed earlier. However, the result of a Wilcoxon significance test, at the conventional significance level of 0.05, indicates that there is no statistically significant difference of predictive performance between HEAD-DT and Auto-WEKA (for both accuracy and GMean measures), for each of the 10 runtime limits.

\begin{figure}[!htpb] \centering 
\resizebox{0.49\columnwidth}{!}{
\subfigure[\normalsize GMean]{
\includegraphics[width=\linewidth]{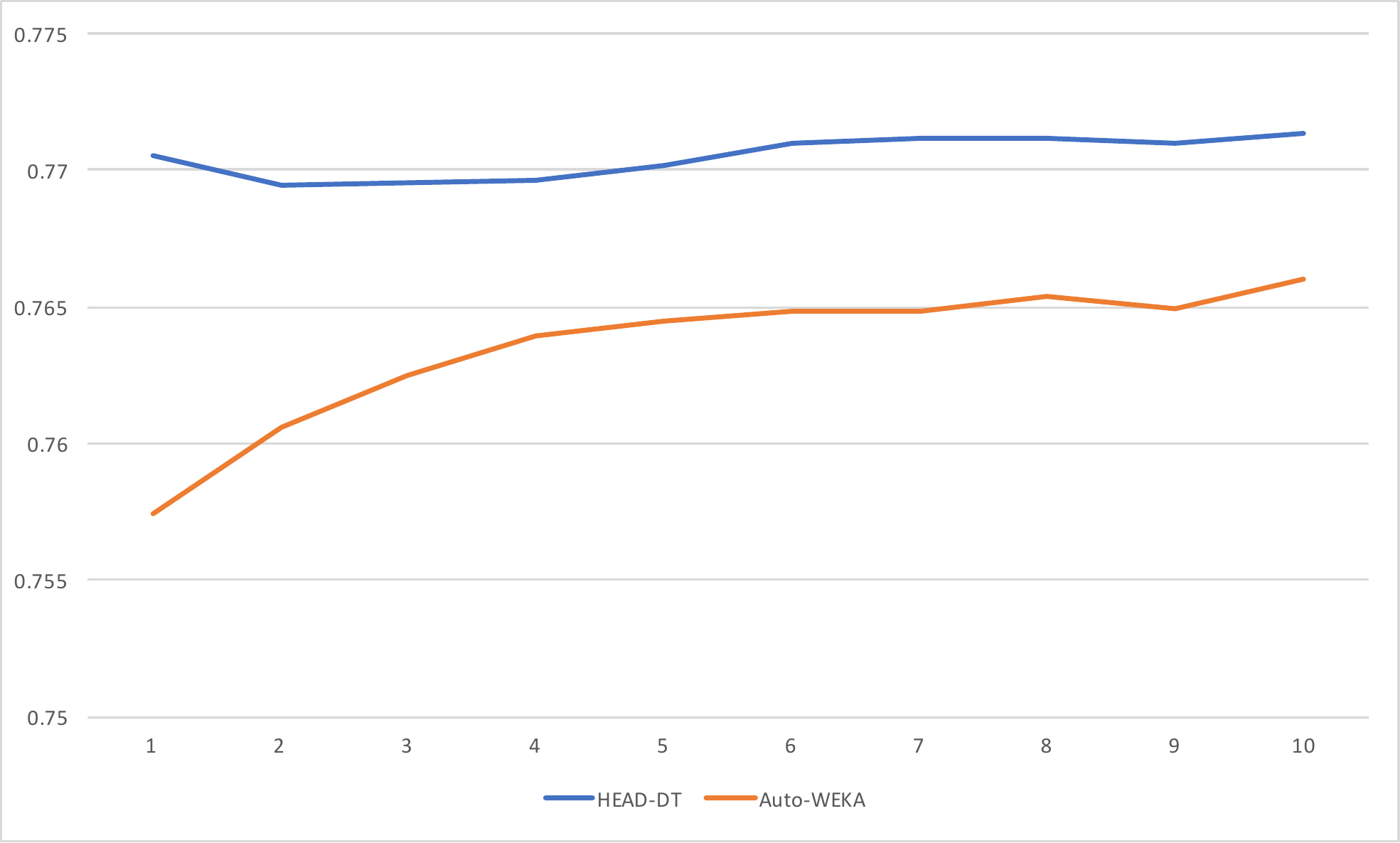} \label{fig:all-ranks-gmean-2-methods}
}}
\resizebox{0.49\columnwidth}{!}{
\subfigure[\normalsize Accuracy]{
\includegraphics[width=\linewidth]{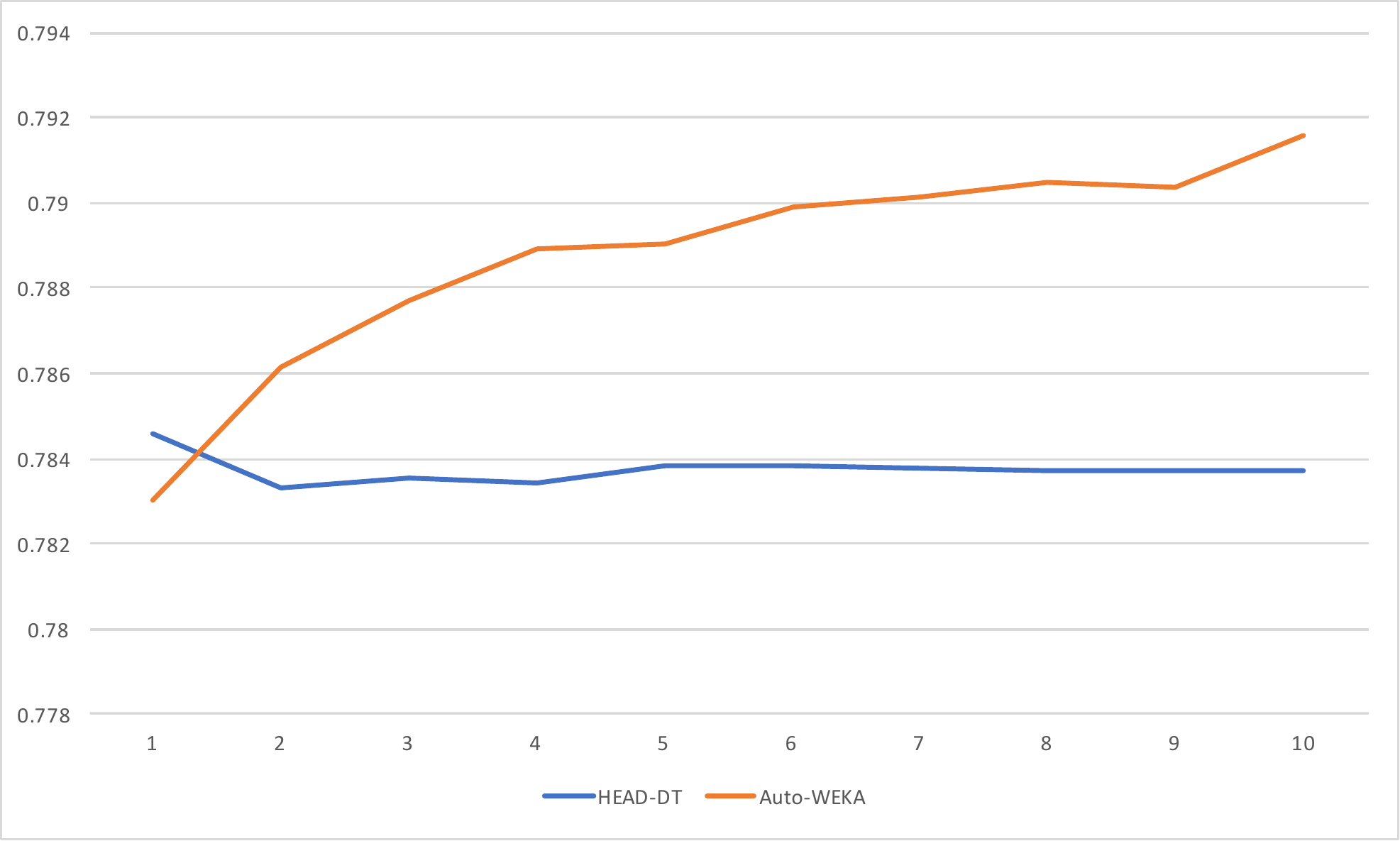} \label{fig:all-ranks-acc-2-methods}
}}
\caption{Evolution of average predictive values for HEAD-DT and Auto-WEKA across the 10 runtime limits.}
\label{fig:all-ranks-2-methods}
\end{figure}

Figure~\ref{fig:all-ranks-gmean-2-methods} shows the evolution of the average GMean values (across all datasets) for Auto-WEKA and HEAD-DT across the 10 runtime limits. This figure shows that HEAD-DT obtains a better (higher) GMean value for all runtime limits.
Figure~\ref{fig:all-ranks-acc-2-methods} shows the same type of evolution for the accuracy measure. In this case, HEAD-DT obtains the best average accuracy for the smallest runtime limit, but Auto-WEKA obtains higher accuracy for all other runtime limits. It should be noted, however, that in both graphs the differences of predictive performance between HEAD-DT and Auto-WEKA are small, less than 1\% in general, across the different runtime limits.

Finally, the presence of signs of overfitting for HEAD-DT and Auto-WEKA was also investigated. This investigation compared HEAD-DT’s GMean values on the
validation set (a holdout part of the training set) and Auto-WEKA-Trees’ GMean values for the
internal 10-CV procedure (on the training set) with their corresponding GMean values on the test
set. A GMean value on the test set much smaller than the corresponding GMean value on the
validation set or internal 10-CV procedure (depending on the method) was considered a sign of
overfitting. This kind of overfitting can be called meta-overfitting, since it occurs at the meta-learning
level, rather than the conventional overfitting at the base-learning level (involving the
difference between GMean values on the training and validation sets).

Note that the meta-overfitting is measured in somewhat different ways in HEAD-DT and Auto-
WEKA due to the different approaches to evaluate candidate solutions during their searches. That
is, HEAD-DT performs a single partition of the training set into two subsets, one for building the
model, the other (validation set) for evaluating the model; whilst Auto-WEKA uses internal cross-validation
on the training set. Despite this difference, the principle used for measuring meta-overfitting
is the same in both types of methods: the degree of meta-overfitting is measured by
comparing predictive performance on the test set (not accessed during the entire execution of HEAD-DT or Auto-WEKA) with the predictive performance on the part of the training set used as a
hold-out set to evaluate the model built from the remaining part of the training set.

For the runtime limit of 1,000s (10,000s), the average GMean value (over all 40 datasets) of HEAD-DT on the validation set is 0.756 (0.758), whilst its average GMean on the test set is 0.770 (0.771). In addition, for the runtime limit of 1,000s (10,000s), the average accuracy value of HEAD-DT on the validation set is 0.766 (0.767), whilst its average accuracy on the test set is 0.785 and 0.784. Hence, HEAD-DT shows no sign of meta-overfitting, since its GMean and accuracy values on the test set are slightly larger than on the validation set. 
This small increase in the GMean and accuracy values on the test set, for both runtime limits, can be explained mainly by two factors. 
First, since the algorithms were evolved by HEAD-DT using the F-measure of precision
and recall in the fitness function, they were not optimizing GMean or
accuracy. Second, the classifier used to classify the test set is in principle a
higher-quality classifier than the one used to classify the validation set, because
the former was induced from all training instances, whilst the latter
was induced from a subset of the training set (excluding the validation set).

Turning to Auto-WEKA, for the runtime limit of 1,000s (10,000s), the average GMean value (over all 40 datasets) for the internal 10-CV of Auto-WEKA is 0.896 (0.899), whilst its GMean on the test set is 0.757 (0.766). In addition, for the runtime limit of 1,000s (10,000s), the average accuracy value for the internal 10-CV of Auto-WEKA is 0.912 (0.915), whilst its accuracy on the test set is 0.783 (0.792). 
Hence, for both the GMean and accuracy measures, Auto-WEKA clearly shows a substantial degree of meta-overfitting.


\section{Conclusions}\label{sec:conclusions}

AutoML is currently a very popular issue, having attracted a great deal of attention, with the proposal of new tools, mainly based on optimization \cite{Hutter2019, Elsken2019, Mohr2018, DasDores2018, Li2018, Larcher2019, Koch2018, Haifeng2019, Fusi2018}. Based on the relevance of AutoML, this work has evaluated four methods for recommending a classification algorithm for a target dataset: three Evolutionary Algorithms (EAs) and Auto-WEKA~\cite{Thornton2013autoweka}, in two sets of experiments.
In the first set of experiments, we have compared the four AutoML methods with the same runtime limit on 20 datasets. Auto-WEKA can recommend classification algorithms of various types (paradigms), whilst each of the three EAs is restricted to recommend a different type of classification algorithm: decision tree, rule induction or Bayesian network classification algorithms, in the case of HEAD-DT, GGP-RI and HHEA-BNC, respectively. In these experiments, there was no statistically significant difference of predictive accuracy between the three best methods, namely two EAs (HEAD-DT and HHEA-BNC) and Auto-WEKA. However, these three methods obtained significantly better predictive accuracy than the other EA (GGP-RI). These results were broadly consistent across the 10 different runtime limits used in the experiments. In the second set of experiments, where a larger set of 40 datasets was used to compare the predictive accuracy of HEAD-DT and Auto-WEKA only, again there was no statistically significant difference between the predictive performance of these two methods. 

However, the focus of HEAD-DT on only on decision-tree algorithms has two advantages from the perspective of other algorithm-evaluation criteria.
First, in applications where it is important that the classification model be interpreted by users (e.g. in medical applications), decision-tree algorithms have the advantage of generating interpretable classification models. By contrast, since Auto-WEKA can select any algorithm out of many types of classification algorithm, it can recommend classification algorithms producing black-box (non-interpretable) models. Indeed, in our experiments, Auto-WEKA often recommended ensembles, which are not easily interpretable. Second, decision-tree algorithms also have the advantage of being in general more scalable to large datasets than several other types of classification algorithms in Auto-WEKA's search space, like neural networks, support vector machines and some ensemble methods.

Overall, when the runtime limit is increased from 1,000s to 10,000s, Auto-WEKA benefits more from the extra search time than HEAD-DT. This seems due to the fact that Auto-WEKA has to explore a much more diverse space of classification algorithms, so it probably requires more time to find the best type of classification algorithm to be recommended for a given input dataset.

In addition, we observed that Auto-WEKA exhibited meta-overfitting, where the GMean values on the training set were substantially lower than the GMean values on the test set, for the best algorithm found by Auto-WEKA. As noted earlier, this meta-overfitting is a form of overfitting at the meta-learning level, due to evaluating many different (base-level) classification algorithms during Auto-WEKA’s search for the best algorithm. This is in contrast to the standard overfitting at the base level, due to evaluating many different models built by the same classification algorithm.

\subsection{Future Work}
It would be interesting to enhance the search process of the EAs by first performing a global search to optimise the candidate algorithms’ (procedural) components, 
followed by a second (global or local) search to optimise the continuous parameters of the best algorithm generated by the first search. Another future research direction is to extend the EAs to produce an ensemble of evolved classification algorithms in a post-processing phase, 
after the EAs have completed their search.

Besides, since Auto-WEKA showed a clear sign of meta-overfitting,
another research direction consists of developing new meta-overfitting-avoidance methods that could potentially improve the predictive performance of Auto-WEKA. Finally, it would be interesting to compare the three EAs and Auto-WEKA to other AutoML methods, such as Auto-sklearn and those described in Section~\ref{related}. This would give us a more detailed assessment about which AutoML method recommends the best classification algorithm, taking into account different datasets.
\appendix
\section{Supplementary Material for: An Extensive Experimental Evaluation of Automated Machine Learning Methods for Recommending Classification Algorithms} \label{sec:appendix}

\setcounter{table}{0}
\renewcommand{\thetable}{A\arabic{table}}

\begin{table}[htbp] \scriptsize
  \centering
  \caption{Results of predictive Specificity values of the baseline methods for building decision tree models.}
\begin{tabular}{lrrrrr}
    \toprule
    data set & \multicolumn{1}{c}{Decision Stump} & \multicolumn{1}{c}{J48} & \multicolumn{1}{c}{LMT} & \multicolumn{1}{c}{RandomTrees} & \multicolumn{1}{c}{REPTree} \\
    \midrule
    CE    & 0.499 & 0.585 & 0.576 & 0.496 & 0.541 \\
    DM    & 0.348 & 0.545 & 0.423 & 0.449 & 0.315 \\
    MM    & 0.312 & 0.496 & 0.403 & 0.484 & 0.292 \\
    SC    & 0.280 & 0.542 & 0.362 & 0.412 & 0.297 \\
    DNA3  & 0.562 & 0.720 & 0.604 & 0.482 & 0.562 \\
    DNA11 & 0.260 & 0.478 & 0.420 & 0.491 & 0.288 \\
    PS    & 0.700 & 0.801 & 0.794 & 0.913 & 0.823 \\
    chen-2002 & 0.857 & 0.798 & 0.944 & 0.826 & 0.756 \\
    chowdary-2006 & 0.965 & 0.951 & 0.969 & 0.918 & 0.930 \\
    nutt-2003-v2 & 0.933 & 0.933 & 0.850 & 0.733 & 0.600 \\
    singh-2002 & 0.746 & 0.781 & 0.901 & 0.754 & 0.696 \\
    west-2001 & 0.862 & 0.815 & 0.878 & 0.578 & 0.782 \\
    dbworld-bodies & 0.770 & 0.743 & 0.819 & 0.686 & 0.758 \\
    dbworld-bodies-stemmed & 0.900 & 0.792 & 0.758 & 0.596 & 0.811 \\
    oh0.wc & 0.857 & 0.972 & 0.979 & 0.925 & 0.962 \\
    oh5.wc & 0.857 & 0.976 & 0.985 & 0.921 & 0.975 \\
    oh10.wc & 0.863 & 0.961 & 0.967 & 0.921 & 0.962 \\
    oh15.wc & 0.851 & 0.966 & 0.973 & 0.921 & 0.952 \\
    re0.wc & 0.736 & 0.915 & 0.940 & 0.881 & 0.909 \\
    re1.wc & 0.829 & 0.970 & 0.979 & 0.941 & 0.970 \\
    \midrule
    Average & 0.699 & 0.787 & 0.776 & 0.716 & 0.709 \\
    \bottomrule
    \end{tabular}%
\end{table}%

\begin{table}[htbp] \scriptsize
  \centering
  \caption{Results of predictive Sensitivity values of the baseline methods for building decision tree models.}
    \begin{tabular}{lrrrrr}
    \toprule
    data set & \multicolumn{1}{c}{Decision Stump} & \multicolumn{1}{c}{J48} & \multicolumn{1}{c}{LMT} & \multicolumn{1}{c}{RandomTrees} & \multicolumn{1}{c}{REPTree} \\
    \midrule
    CE    & 0.623 & 0.648 & 0.659 & 0.567 & 0.642 \\
    DM    & 0.614 & 0.664 & 0.623 & 0.555 & 0.623 \\
    MM    & 0.631 & 0.661 & 0.640 & 0.631 & 0.708 \\
    SC    & 0.827 & 0.822 & 0.835 & 0.774 & 0.807 \\
    DNA3  & 0.863 & 0.841 & 0.848 & 0.733 & 0.863 \\
    DNA11 & 0.720 & 0.777 & 0.740 & 0.706 & 0.726 \\
    PS    & 0.981 & 0.985 & 0.985 & 0.985 & 0.983 \\
    chen-2002 & 0.821 & 0.810 & 0.949 & 0.827 & 0.787 \\
    chowdary-2006 & 0.971 & 0.951 & 0.972 & 0.923 & 0.944 \\
    nutt-2003-v2 & 0.867 & 0.867 & 0.850 & 0.717 & 0.450 \\
    singh-2002 & 0.744 & 0.783 & 0.903 & 0.756 & 0.694 \\
    west-2001 & 0.855 & 0.835 & 0.855 & 0.555 & 0.735 \\
    dbworld-bodies & 0.738 & 0.740 & 0.814 & 0.698 & 0.733 \\
    dbworld-bodies-stemmed & 0.867 & 0.783 & 0.767 & 0.612 & 0.798 \\
    oh0.wc & 0.351 & 0.820 & 0.874 & 0.463 & 0.779 \\
    oh5.wc & 0.261 & 0.807 & 0.883 & 0.356 & 0.805 \\
    oh10.wc & 0.262 & 0.727 & 0.754 & 0.368 & 0.742 \\
    oh15.wc & 0.272 & 0.736 & 0.810 & 0.387 & 0.653 \\
    re0.wc & 0.537 & 0.745 & 0.832 & 0.559 & 0.739 \\
    re1.wc & 0.396 & 0.798 & 0.859 & 0.472 & 0.786 \\
    \midrule
    Average & 0.660 & 0.790 & 0.823 & 0.632 & 0.750 \\
    \bottomrule
    \end{tabular}%
\end{table}%

\begin{table}[htbp] \scriptsize
  \centering
  \caption{Results of predictive GMean values of the baseline methods for building decision tree models.}
    \begin{tabular}{lrrrrr}
    \toprule
    data set & \multicolumn{1}{c}{Decision Stump} & \multicolumn{1}{c}{J48} & \multicolumn{1}{c}{LMT} & \multicolumn{1}{c}{RandomTrees} & \multicolumn{1}{c}{REPTree} \\
    \midrule
    CE    & 0.557 & 0.616 & 0.616 & 0.530 & 0.589 \\
    DM    & 0.460 & 0.600 & 0.511 & 0.496 & 0.442 \\
    MM    & 0.435 & 0.569 & 0.499 & 0.537 & 0.451 \\
    SC    & 0.464 & 0.663 & 0.535 & 0.556 & 0.466 \\
    DNA3  & 0.694 & 0.773 & 0.707 & 0.584 & 0.694 \\
    DNA11 & 0.426 & 0.601 & 0.553 & 0.586 & 0.450 \\
    PS    & 0.828 & 0.888 & 0.884 & 0.948 & 0.899 \\
    chen-2002 & 0.838 & 0.804 & 0.947 & 0.826 & 0.771 \\
    chowdary-2006 & 0.968 & 0.951 & 0.970 & 0.920 & 0.937 \\
    nutt-2003-v2 & 0.898 & 0.898 & 0.846 & 0.721 & 0.507 \\
    singh-2002 & 0.745 & 0.782 & 0.902 & 0.755 & 0.695 \\
    west-2001 & 0.858 & 0.823 & 0.866 & 0.563 & 0.756 \\
    dbworld-bodies & 0.753 & 0.741 & 0.816 & 0.691 & 0.745 \\
    dbworld-bodies-stemmed & 0.883 & 0.787 & 0.762 & 0.603 & 0.804 \\
    oh0.wc & 0.548 & 0.892 & 0.925 & 0.654 & 0.865 \\
    oh5.wc & 0.473 & 0.887 & 0.933 & 0.568 & 0.886 \\
    oh10.wc & 0.475 & 0.835 & 0.854 & 0.581 & 0.844 \\
    oh15.wc & 0.480 & 0.842 & 0.888 & 0.595 & 0.788 \\
    re0.wc & 0.629 & 0.826 & 0.884 & 0.702 & 0.820 \\
    re1.wc & 0.573 & 0.880 & 0.917 & 0.666 & 0.873 \\
    \midrule
    Average & 0.649 & 0.783 & 0.791 & 0.654 & 0.714 \\
    \bottomrule
    \end{tabular}%
\end{table}%

\begin{table}[htbp] \scriptsize
  \centering
  \caption{Results of predictive Specificity values of HEAD-DT for timeouts from 1,000s to 5,000s.}
    \begin{tabular}{lrrrrr}
    \toprule
    data set & \multicolumn{1}{c}{1,000s} & \multicolumn{1}{c}{2,000s} & \multicolumn{1}{c}{3,000s} & \multicolumn{1}{c}{4,000s} & \multicolumn{1}{c}{5,000s} \\
    \midrule
    CE    & 0.693 & 0.664 & 0.656 & 0.691 & 0.676 \\
    DM    & 0.808 & 0.758 & 0.758 & 0.758 & 0.825 \\
    MM    & 0.733 & 0.739 & 0.739 & 0.739 & 0.675 \\
    SC    & 0.767 & 0.724 & 0.701 & 0.701 & 0.701 \\
    DNA3 & 0.846 & 0.846 & 0.846 & 0.846 & 0.846 \\
    DNA11 & 0.854 & 0.833 & 0.833 & 0.833 & 0.833 \\
    PS    & 0.978 & 0.989 & 0.989 & 0.975 & 0.978 \\
    chen-2002 & 0.964 & 0.937 & 0.937 & 0.937 & 0.937 \\
    chowdary-2006 & 0.965 & 0.965 & 0.965 & 0.965 & 0.965 \\
    nutt-2003-v2 & 0.983 & 0.983 & 0.983 & 0.983 & 0.983 \\
    singh-2002 & 0.852 & 0.852 & 0.852 & 0.852 & 0.852 \\
    west-2001 & 0.973 & 0.973 & 0.973 & 0.973 & 0.973 \\
    dbworld-bodies & 0.844 & 0.844 & 0.844 & 0.844 & 0.844 \\
    dbworld-bodies-stemmed & 0.846 & 0.846 & 0.846 & 0.846 & 0.846 \\
    oh0.wc & 0.984 & 0.984 & 0.981 & 0.980 & 0.980 \\
    oh5.wc & 0.991 & 0.991 & 0.990 & 0.991 & 0.990 \\
    oh10.wc & 0.975 & 0.975 & 0.975 & 0.975 & 0.979 \\
    oh15.wc & 0.977 & 0.978 & 0.979 & 0.980 & 0.980 \\
    re0.wc & 0.961 & 0.952 & 0.953 & 0.954 & 0.955 \\
    re1.wc & 0.986 & 0.985 & 0.985 & 0.983 & 0.983 \\
    \midrule
    Average & 0.899 & 0.891 & 0.889 & 0.890 & 0.890 \\
    \bottomrule
    \end{tabular}%
\end{table}%

\newpage

\begin{table}[htbp] \scriptsize
  \centering
  \caption{Results of predictive Specificity values of HEAD-DT for timeouts from 6,000s to 10,000s.}
    \begin{tabular}{lrrrrr}
    \toprule
    data set & \multicolumn{1}{c}{6,000s} & \multicolumn{1}{c}{7,000s} & \multicolumn{1}{c}{8,000s} & \multicolumn{1}{c}{9,000s} & \multicolumn{1}{c}{10,000s} \\
    \midrule
    CE    & 0.676 & 0.676 & 0.682 & 0.708 & 0.708 \\
    DM    & 0.825 & 0.796 & 0.796 & 0.796 & 0.796 \\
    MM    & 0.675 & 0.675 & 0.675 & 0.675 & 0.675 \\
    SC    & 0.701 & 0.701 & 0.722 & 0.722 & 0.722 \\
    DNA3 & 0.846 & 0.846 & 0.846 & 0.846 & 0.846 \\
    DNA11 & 0.833 & 0.807 & 0.807 & 0.807 & 0.807 \\
    PS    & 0.978 & 0.978 & 0.978 & 0.978 & 0.978 \\
    chen-2002 & 0.937 & 0.937 & 0.937 & 0.937 & 0.937 \\
    chowdary-2006 & 0.965 & 0.965 & 0.965 & 0.965 & 0.965 \\
    nutt-2003-v2 & 0.983 & 0.983 & 0.983 & 0.983 & 0.983 \\
    singh-2002 & 0.852 & 0.852 & 0.852 & 0.852 & 0.852 \\
    west-2001 & 0.973 & 0.973 & 0.973 & 0.973 & 0.973 \\
    dbworld-bodies & 0.844 & 0.844 & 0.844 & 0.844 & 0.844 \\
    dbworld-bodies-stemmed & 0.846 & 0.846 & 0.846 & 0.846 & 0.846 \\
    oh0.wc & 0.981 & 0.982 & 0.982 & 0.982 & 0.982 \\
    oh5.wc & 0.990 & 0.990 & 0.990 & 0.990 & 0.990 \\
    oh10.wc & 0.979 & 0.979 & 0.980 & 0.980 & 0.980 \\
    oh15.wc & 0.982 & 0.982 & 0.982 & 0.982 & 0.982 \\
    re0.wc & 0.954 & 0.957 & 0.957 & 0.957 & 0.957 \\
    re1.wc & 0.984 & 0.984 & 0.985 & 0.985 & 0.985 \\
    \midrule
    Average & 0.890 & 0.888 & 0.889 & 0.890 & 0.890 \\
    \bottomrule
    \end{tabular}%
\end{table}%

\begin{table}[htbp] \scriptsize
  \centering
  \caption{Results of predictive Sensitivity values of HEAD-DT for timeouts from 1,000s to 5,000s.}
    \begin{tabular}{lrrrrr}
    \toprule
    data set & \multicolumn{1}{c}{1,000s} & \multicolumn{1}{c}{2,000s} & \multicolumn{1}{c}{3,000s} & \multicolumn{1}{c}{4,000s} & \multicolumn{1}{c}{5,000s} \\
    \midrule
    CE    & 0.759 & 0.749 & 0.736 & 0.759 & 0.751 \\
    DM    & 0.867 & 0.842 & 0.842 & 0.842 & 0.875 \\
    MM    & 0.867 & 0.878 & 0.878 & 0.878 & 0.875 \\
    SC    & 0.943 & 0.927 & 0.915 & 0.915 & 0.915 \\
    DNA3 & 0.935 & 0.935 & 0.935 & 0.935 & 0.935 \\
    DNA11 & 0.933 & 0.919 & 0.919 & 0.919 & 0.919 \\
    PS    & 0.998 & 0.999 & 0.999 & 0.998 & 0.998 \\
    chen-2002 & 0.966 & 0.950 & 0.950 & 0.950 & 0.950 \\
    chowdary-2006 & 0.971 & 0.971 & 0.971 & 0.971 & 0.971 \\
    nutt-2003-v2 & 0.967 & 0.967 & 0.967 & 0.967 & 0.967 \\
    singh-2002 & 0.851 & 0.851 & 0.851 & 0.851 & 0.851 \\
    west-2001 & 0.960 & 0.960 & 0.960 & 0.960 & 0.960 \\
    dbworld-bodies & 0.831 & 0.831 & 0.831 & 0.831 & 0.831 \\
    dbworld-bodies-stemmed & 0.829 & 0.829 & 0.829 & 0.829 & 0.829 \\
    oh0.wc & 0.896 & 0.901 & 0.887 & 0.884 & 0.884 \\
    oh5.wc & 0.930 & 0.925 & 0.922 & 0.922 & 0.917 \\
    oh10.wc & 0.834 & 0.839 & 0.840 & 0.837 & 0.855 \\
    oh15.wc & 0.845 & 0.851 & 0.855 & 0.858 & 0.859 \\
    re0.wc & 0.888 & 0.870 & 0.874 & 0.878 & 0.878 \\
    re1.wc & 0.893 & 0.893 & 0.887 & 0.874 & 0.879 \\
    \midrule
    Average & 0.898 & 0.894 & 0.892 & 0.893 & 0.895 \\
    \bottomrule
    \end{tabular}%
\end{table}%

\begin{table}[htbp] \scriptsize
  \centering
  \caption{Results of predictive Sensitivity values of HEAD-DT for timeouts from 6,000s to 10,000s.}
    \begin{tabular}{lrrrrr}
    \toprule
    data set & \multicolumn{1}{c}{6,000s} & \multicolumn{1}{c}{7,000s} & \multicolumn{1}{c}{8,000s} & \multicolumn{1}{c}{9,000s} & \multicolumn{1}{c}{10,000s} \\
    \midrule
    CE    & 0.751 & 0.751 & 0.764 & 0.779 & 0.779 \\
    DM    & 0.875 & 0.867 & 0.867 & 0.867 & 0.867 \\
    MM    & 0.875 & 0.875 & 0.875 & 0.875 & 0.875 \\
    SC    & 0.915 & 0.911 & 0.919 & 0.919 & 0.915 \\
    DNA3 & 0.935 & 0.935 & 0.935 & 0.935 & 0.935 \\
    DNA11 & 0.919 & 0.911 & 0.911 & 0.911 & 0.911 \\
    PS    & 0.998 & 0.998 & 0.998 & 0.998 & 0.998 \\
    chen-2002 & 0.950 & 0.950 & 0.950 & 0.950 & 0.950 \\
    chowdary-2006 & 0.971 & 0.971 & 0.971 & 0.971 & 0.971 \\
    nutt-2003-v2 & 0.967 & 0.967 & 0.967 & 0.967 & 0.967 \\
    singh-2002 & 0.851 & 0.851 & 0.851 & 0.851 & 0.851 \\
    west-2001 & 0.960 & 0.960 & 0.960 & 0.960 & 0.960 \\
    dbworld-bodies & 0.831 & 0.831 & 0.831 & 0.831 & 0.831 \\
    dbworld-bodies-stemmed & 0.829 & 0.829 & 0.829 & 0.829 & 0.829 \\
    oh0.wc & 0.886 & 0.892 & 0.892 & 0.892 & 0.892 \\
    oh5.wc & 0.912 & 0.912 & 0.912 & 0.912 & 0.916 \\
    oh10.wc & 0.855 & 0.855 & 0.858 & 0.858 & 0.858 \\
    oh15.wc & 0.865 & 0.865 & 0.874 & 0.872 & 0.872 \\
    re0.wc & 0.875 & 0.880 & 0.880 & 0.880 & 0.880 \\
    re1.wc & 0.883 & 0.891 & 0.893 & 0.893 & 0.893 \\
    \midrule
    Average & 0.895 & 0.895 & 0.897 & 0.897 & 0.897 \\
    \bottomrule
    \end{tabular}%
\end{table}%

\begin{table}[htbp] \scriptsize
  \centering
  \caption{Results of predictive GMean values of HEAD-DT for timeouts from 1,000s to 5,000s.}
    \begin{tabular}{lrrrrr}
    \toprule
    data set & \multicolumn{1}{c}{1,000s} & \multicolumn{1}{c}{2,000s} & \multicolumn{1}{c}{3,000s} & \multicolumn{1}{c}{4,000s} & \multicolumn{1}{c}{5,000s} \\
    \midrule
    CE    & 0.725 & 0.705 & 0.694 & 0.724 & 0.712 \\
    DM    & 0.834 & 0.794 & 0.794 & 0.794 & 0.846 \\
    MM    & 0.789 & 0.797 & 0.797 & 0.797 & 0.756 \\
    SC    & 0.846 & 0.806 & 0.788 & 0.788 & 0.788 \\
    DNA3  & 0.887 & 0.887 & 0.887 & 0.887 & 0.887 \\
    DNA11 & 0.890 & 0.873 & 0.873 & 0.873 & 0.873 \\
    PS    & 0.988 & 0.994 & 0.994 & 0.986 & 0.988 \\
    chen-2002 & 0.965 & 0.943 & 0.943 & 0.943 & 0.943 \\
    chowdary-2006 & 0.968 & 0.968 & 0.968 & 0.968 & 0.968 \\
    nutt-2003-v2 & 0.975 & 0.975 & 0.975 & 0.975 & 0.975 \\
    singh-2002 & 0.852 & 0.852 & 0.852 & 0.852 & 0.852 \\
    west-2001 & 0.967 & 0.967 & 0.967 & 0.967 & 0.967 \\
    dbworld-bodies & 0.837 & 0.837 & 0.837 & 0.837 & 0.837 \\
    dbworld-bodies-stemmed & 0.837 & 0.837 & 0.837 & 0.837 & 0.837 \\
    oh0.wc & 0.939 & 0.941 & 0.933 & 0.931 & 0.931 \\
    oh5.wc & 0.960 & 0.957 & 0.955 & 0.955 & 0.953 \\
    oh10.wc & 0.901 & 0.904 & 0.904 & 0.902 & 0.915 \\
    oh15.wc & 0.909 & 0.912 & 0.915 & 0.916 & 0.917 \\
    re0.wc & 0.924 & 0.910 & 0.913 & 0.915 & 0.916 \\
    re1.wc & 0.938 & 0.938 & 0.934 & 0.927 & 0.929 \\
    \midrule
    Average & 0.896 & 0.890 & 0.888 & 0.889 & 0.889 \\
    \bottomrule
    \end{tabular}%
\end{table}%

\begin{table}[htbp] \scriptsize
  \centering
  \caption{Results of predictive GMean values of HEAD-DT for timeouts from 6,000s to 10,000s.}
    \begin{tabular}{lrrrrr}
    \toprule
    data set & \multicolumn{1}{c}{6,000s} & \multicolumn{1}{c}{7,000s} & \multicolumn{1}{c}{8,000s} & \multicolumn{1}{c}{9,000s} & \multicolumn{1}{c}{10,000s} \\
    \midrule
    CE    & 0.712 & 0.712 & 0.721 & 0.742 & 0.742 \\
    DM    & 0.846 & 0.827 & 0.827 & 0.827 & 0.827 \\
    MM    & 0.756 & 0.756 & 0.756 & 0.756 & 0.756 \\
    SC    & 0.788 & 0.786 & 0.803 & 0.803 & 0.801 \\
    DNA3  & 0.887 & 0.887 & 0.887 & 0.887 & 0.887 \\
    DNA11 & 0.873 & 0.855 & 0.855 & 0.855 & 0.855 \\
    PS    & 0.988 & 0.988 & 0.988 & 0.988 & 0.988 \\
    chen-2002 & 0.943 & 0.943 & 0.943 & 0.943 & 0.943 \\
    chowdary-2006 & 0.968 & 0.968 & 0.968 & 0.968 & 0.968 \\
    nutt-2003-v2 & 0.975 & 0.975 & 0.975 & 0.975 & 0.975 \\
    singh-2002 & 0.852 & 0.852 & 0.852 & 0.852 & 0.852 \\
    west-2001 & 0.967 & 0.967 & 0.967 & 0.967 & 0.967 \\
    dbworld-bodies & 0.837 & 0.837 & 0.837 & 0.837 & 0.837 \\
    dbworld-bodies-stemmed & 0.837 & 0.837 & 0.837 & 0.837 & 0.837 \\
    oh0.wc & 0.932 & 0.936 & 0.936 & 0.936 & 0.936 \\
    oh5.wc & 0.950 & 0.950 & 0.950 & 0.950 & 0.952 \\
    oh10.wc & 0.915 & 0.915 & 0.917 & 0.917 & 0.917 \\
    oh15.wc & 0.921 & 0.921 & 0.926 & 0.925 & 0.925 \\
    re0.wc & 0.913 & 0.917 & 0.917 & 0.917 & 0.917 \\
    re1.wc & 0.932 & 0.936 & 0.938 & 0.938 & 0.938 \\
    \midrule
    Average & 0.890 & 0.888 & 0.890 & 0.891 & 0.891 \\
    \bottomrule
    \end{tabular}%
\end{table}%

\begin{table}[htbp] \scriptsize
  \centering
  \caption{Results of predictive Specificity values of Auto-WEKA-Trees for timeouts from 1,000s to 5,000s.}
    \begin{tabular}{lrrrrr}
    \toprule
    data set & \multicolumn{1}{c}{1,000s} & \multicolumn{1}{c}{2,000s} & \multicolumn{1}{c}{3,000s} & \multicolumn{1}{c}{4,000s} & \multicolumn{1}{c}{5,000s} \\
    \midrule
    CE    & 0.539 & 0.547 & 0.549 & 0.552 & 0.552 \\
    DM    & 0.378 & 0.363 & 0.361 & 0.366 & 0.388 \\
    MM    & 0.399 & 0.418 & 0.417 & 0.418 & 0.418 \\
    SC    & 0.254 & 0.253 & 0.253 & 0.259 & 0.254 \\
    DNA3 & 0.554 & 0.593 & 0.578 & 0.589 & 0.612 \\
    DNA11 & 0.403 & 0.376 & 0.401 & 0.377 & 0.362 \\
    PS    & 0.699 & 0.702 & 0.711 & 0.721 & 0.722 \\
    chen-2002 & 0.924 & 0.926 & 0.914 & 0.917 & 0.918 \\
    chowdary-2006 & 0.993 & 0.994 & 0.994 & 0.994 & 0.993 \\
    nutt-2003-v2 & 0.879 & 0.854 & 0.854 & 0.854 & 0.854 \\
    singh-2002 & 0.892 & 0.886 & 0.886 & 0.881 & 0.875 \\
    west-2001 & 0.914 & 0.914 & 0.914 & 0.914 & 0.914 \\
    dbworld-bodies & 0.738 & 0.743 & 0.734 & 0.738 & 0.763 \\
    dbworld-bodies-stemmed & 0.863 & 0.843 & 0.793 & 0.769 & 0.771 \\
    oh0.wc & 0.965 & 0.965 & 0.966 & 0.966 & 0.966 \\
    oh5.wc & 0.981 & 0.981 & 0.980 & 0.980 & 0.980 \\
    oh10.wc & 0.959 & 0.959 & 0.959 & 0.958 & 0.959 \\
    oh15.wc & 0.967 & 0.967 & 0.968 & 0.968 & 0.969 \\
    re0.wc & 0.921 & 0.919 & 0.920 & 0.920 & 0.920 \\
    re1.wc & 0.968 & 0.968 & 0.969 & 0.969 & 0.969 \\
    \midrule
    Average & 0.759 & 0.759 & 0.756 & 0.756 & 0.758 \\
    \bottomrule
    \end{tabular}%
\end{table}%

\begin{table}[htbp] \scriptsize
  \centering
  \caption{Results of predictive Specificity values of Auto-WEKA-Trees for timeouts from 6,000s to 10,000s.}
    \begin{tabular}{lrrrrr}
    \toprule
    data set & \multicolumn{1}{c}{6,000s} & \multicolumn{1}{c}{7,000s} & \multicolumn{1}{c}{8,000s} & \multicolumn{1}{c}{9,000s} & \multicolumn{1}{c}{10,000s} \\
    \midrule
    CE    & 0.553 & 0.558 & 0.553 & 0.555 & 0.558 \\
    DM    & 0.387 & 0.389 & 0.388 & 0.375 & 0.387 \\
    MM    & 0.418 & 0.414 & 0.413 & 0.416 & 0.402 \\
    SC    & 0.259 & 0.254 & 0.290 & 0.258 & 0.269 \\
    DNA3  & 0.600 & 0.610 & 0.598 & 0.608 & 0.601 \\
    DNA11 & 0.388 & 0.379 & 0.378 & 0.422 & 0.387 \\
    PS    & 0.723 & 0.744 & 0.736 & 0.748 & 0.747 \\
    chen-2002 & 0.920 & 0.923 & 0.915 & 0.914 & 0.912 \\
    chowdary-2006 & 0.993 & 0.993 & 0.993 & 0.993 & 0.988 \\
    nutt-2003-v2 & 0.858 & 0.858 & 0.850 & 0.854 & 0.875 \\
    singh-2002 & 0.872 & 0.875 & 0.872 & 0.872 & 0.877 \\
    west-2001 & 0.914 & 0.908 & 0.908 & 0.908 & 0.908 \\
    dbworld-bodies & 0.737 & 0.739 & 0.739 & 0.739 & 0.739 \\
    dbworld-bodies-stemmed & 0.769 & 0.769 & 0.772 & 0.772 & 0.772 \\
    oh0.wc & 0.966 & 0.966 & 0.966 & 0.966 & 0.966 \\
    oh5.wc & 0.981 & 0.980 & 0.980 & 0.980 & 0.980 \\
    oh10.wc & 0.958 & 0.959 & 0.958 & 0.959 & 0.958 \\
    oh15.wc & 0.969 & 0.969 & 0.969 & 0.969 & 0.969 \\
    re0.wc & 0.920 & 0.919 & 0.917 & 0.917 & 0.917 \\
    re1.wc & 0.969 & 0.969 & 0.969 & 0.969 & 0.969 \\
    \midrule
    Average & 0.758 & 0.759 & 0.758 & 0.760 & 0.759 \\
    \bottomrule
    \end{tabular}%
\end{table}%

\begin{table}[htbp] \scriptsize
  \centering
  \caption{Results of predictive Sensitivity values of Auto-WEKA-Trees for timeouts from 1,000s to 5,000s.}
    \begin{tabular}{lrrrrr}
    \toprule
    data set & \multicolumn{1}{c}{1,000s} & \multicolumn{1}{c}{2,000s} & \multicolumn{1}{c}{3,000s} & \multicolumn{1}{c}{4,000s} & \multicolumn{1}{c}{5,000s} \\
    \midrule
    CE    & 0.635 & 0.643 & 0.644 & 0.645 & 0.645 \\
    DM    & 0.634 & 0.628 & 0.630 & 0.628 & 0.641 \\
    MM    & 0.683 & 0.692 & 0.686 & 0.692 & 0.692 \\
    SC    & 0.820 & 0.817 & 0.817 & 0.818 & 0.820 \\
    DNA3  & 0.833 & 0.847 & 0.831 & 0.834 & 0.843 \\
    DNA11 & 0.704 & 0.715 & 0.720 & 0.706 & 0.708 \\
    PS    & 0.976 & 0.976 & 0.976 & 0.977 & 0.977 \\
    chen-2002 & 0.931 & 0.934 & 0.926 & 0.927 & 0.927 \\
    chowdary-2006 & 0.991 & 0.993 & 0.993 & 0.993 & 0.991 \\
    nutt-2003-v2 & 0.821 & 0.796 & 0.796 & 0.796 & 0.796 \\
    singh-2002 & 0.893 & 0.889 & 0.889 & 0.884 & 0.877 \\
    west-2001 & 0.899 & 0.899 & 0.899 & 0.899 & 0.899 \\
    dbworld-bodies & 0.748 & 0.747 & 0.739 & 0.745 & 0.770 \\
    dbworld-bodies-stemmed & 0.848 & 0.830 & 0.786 & 0.770 & 0.770 \\
    oh0.wc & 0.802 & 0.804 & 0.808 & 0.810 & 0.809 \\
    oh5.wc & 0.807 & 0.807 & 0.805 & 0.807 & 0.807 \\
    oh10.wc & 0.725 & 0.721 & 0.726 & 0.721 & 0.724 \\
    oh15.wc & 0.775 & 0.778 & 0.783 & 0.783 & 0.788 \\
    re0.wc & 0.745 & 0.768 & 0.770 & 0.769 & 0.768 \\
    re1.wc & 0.766 & 0.766 & 0.767 & 0.767 & 0.768 \\
    \midrule
    Average & 0.802 & 0.803 & 0.800 & 0.799 & 0.801 \\
    \bottomrule
    \end{tabular}%
\end{table}%

\begin{table}[htbp] \scriptsize
  \centering
  \caption{Results of predictive Sensitivity values of Auto-WEKA-Trees for timeouts from 6,000s to 10,000s.}
    \begin{tabular}{lrrrrr}
    \toprule
    data set & \multicolumn{1}{c}{6,000s} & \multicolumn{1}{c}{7,000s} & \multicolumn{1}{c}{8,000s} & \multicolumn{1}{c}{9,000s} & \multicolumn{1}{c}{10,000s} \\
    \midrule
    CE    & 0.645 & 0.648 & 0.645 & 0.646 & 0.648 \\
    DM    & 0.645 & 0.642 & 0.640 & 0.641 & 0.660 \\
    MM    & 0.692 & 0.692 & 0.689 & 0.686 & 0.683 \\
    SC    & 0.818 & 0.820 & 0.825 & 0.817 & 0.818 \\
    DNA3  & 0.842 & 0.843 & 0.843 & 0.838 & 0.834 \\
    DNA11 & 0.712 & 0.711 & 0.717 & 0.721 & 0.715 \\
    PS    & 0.977 & 0.978 & 0.978 & 0.979 & 0.978 \\
    chen-2002 & 0.930 & 0.931 & 0.926 & 0.926 & 0.926 \\
    chowdary-2006 & 0.991 & 0.991 & 0.991 & 0.991 & 0.983 \\
    nutt-2003-v2 & 0.804 & 0.804 & 0.788 & 0.796 & 0.813 \\
    singh-2002 & 0.874 & 0.877 & 0.874 & 0.874 & 0.879 \\
    west-2001 & 0.899 & 0.893 & 0.893 & 0.893 & 0.893 \\
    dbworld-bodies & 0.754 & 0.755 & 0.755 & 0.755 & 0.755 \\
    dbworld-bodies-stemmed & 0.766 & 0.766 & 0.770 & 0.770 & 0.770 \\
    oh0.wc & 0.810 & 0.810 & 0.810 & 0.805 & 0.806 \\
    oh5.wc & 0.810 & 0.808 & 0.808 & 0.808 & 0.810 \\
    oh10.wc & 0.721 & 0.723 & 0.727 & 0.728 & 0.722 \\
    oh15.wc & 0.790 & 0.791 & 0.792 & 0.791 & 0.792 \\
    re0.wc & 0.769 & 0.768 & 0.768 & 0.767 & 0.770 \\
    re1.wc & 0.768 & 0.768 & 0.768 & 0.767 & 0.769 \\
    \midrule
    Average & 0.801 & 0.801 & 0.800 & 0.800 & 0.801 \\
    \bottomrule
    \end{tabular}%
\end{table}%

\begin{table}[htbp] \scriptsize
  \centering
  \caption{Results of predictive GMean values of Auto-WEKA-Trees for timeouts from 1,000s to 5,000s.}
    \begin{tabular}{lrrrrr}
    \toprule
    data set & \multicolumn{1}{c}{1,000s} & \multicolumn{1}{c}{2,000s} & \multicolumn{1}{c}{3,000s} & \multicolumn{1}{c}{4,000s} & \multicolumn{1}{c}{5,000s} \\
    \midrule
    CE    & 0.539 & 0.547 & 0.549 & 0.552 & 0.552 \\
    DM    & 0.378 & 0.363 & 0.361 & 0.366 & 0.388 \\
    MM    & 0.399 & 0.418 & 0.417 & 0.418 & 0.418 \\
    SC    & 0.254 & 0.253 & 0.253 & 0.259 & 0.254 \\
    DNA3  & 0.554 & 0.593 & 0.578 & 0.589 & 0.612 \\
    DNA11 & 0.403 & 0.376 & 0.401 & 0.377 & 0.362 \\
    PS    & 0.699 & 0.702 & 0.711 & 0.721 & 0.722 \\
    chen-2002 & 0.924 & 0.926 & 0.914 & 0.917 & 0.918 \\
    chowdary-2006 & 0.993 & 0.994 & 0.994 & 0.994 & 0.993 \\
    nutt-2003-v2 & 0.879 & 0.854 & 0.854 & 0.854 & 0.854 \\
    singh-2002 & 0.892 & 0.886 & 0.886 & 0.881 & 0.875 \\
    west-2001 & 0.914 & 0.914 & 0.914 & 0.914 & 0.914 \\
    dbworld-bodies & 0.738 & 0.743 & 0.734 & 0.738 & 0.763 \\
    dbworld-bodies-stemmed & 0.863 & 0.843 & 0.793 & 0.769 & 0.771 \\
    oh0.wc & 0.965 & 0.965 & 0.966 & 0.966 & 0.966 \\
    oh5.wc & 0.981 & 0.981 & 0.980 & 0.980 & 0.980 \\
    oh10.wc & 0.959 & 0.959 & 0.959 & 0.958 & 0.959 \\
    oh15.wc & 0.967 & 0.967 & 0.968 & 0.968 & 0.969 \\
    re0.wc & 0.921 & 0.919 & 0.920 & 0.920 & 0.920 \\
    re1.wc & 0.968 & 0.968 & 0.969 & 0.969 & 0.969 \\
    \midrule
    Average & 0.759 & 0.759 & 0.756 & 0.756 & 0.758 \\
    \bottomrule
    \end{tabular}%
  \label{tab:AW-Trees-timeoutsA}%
\end{table}%

\begin{table}[htbp] \scriptsize
  \centering
  \caption{Results of predictive GMean values of Auto-WEKA-Trees for timeouts from 6,000s to 10,000s.}
    \begin{tabular}{lrrrrr}
    \toprule
    data set & \multicolumn{1}{c}{6,000s} & \multicolumn{1}{c}{7,000s} & \multicolumn{1}{c}{8,000s} & \multicolumn{1}{c}{9,000s} & \multicolumn{1}{c}{10,000s} \\
    \midrule
    CE    & 0.553 & 0.558 & 0.553 & 0.555 & 0.558 \\
    DM    & 0.387 & 0.389 & 0.388 & 0.375 & 0.387 \\
    MM    & 0.418 & 0.414 & 0.413 & 0.416 & 0.402 \\
    SC    & 0.259 & 0.254 & 0.290 & 0.258 & 0.269 \\
    DNA3  & 0.600 & 0.610 & 0.598 & 0.608 & 0.601 \\
    DNA11 & 0.388 & 0.379 & 0.378 & 0.422 & 0.387 \\
    PS    & 0.723 & 0.744 & 0.736 & 0.748 & 0.747 \\
    chen-2002 & 0.920 & 0.923 & 0.915 & 0.914 & 0.912 \\
    chowdary-2006 & 0.993 & 0.993 & 0.993 & 0.993 & 0.988 \\
    nutt-2003-v2 & 0.858 & 0.858 & 0.850 & 0.854 & 0.875 \\
    singh-2002 & 0.872 & 0.875 & 0.872 & 0.872 & 0.877 \\
    west-2001 & 0.914 & 0.908 & 0.908 & 0.908 & 0.908 \\
    dbworld-bodies & 0.737 & 0.739 & 0.739 & 0.739 & 0.739 \\
    dbworld-bodies-stemmed & 0.769 & 0.769 & 0.772 & 0.772 & 0.772 \\
    oh0.wc & 0.966 & 0.966 & 0.966 & 0.966 & 0.966 \\
    oh5.wc & 0.981 & 0.980 & 0.980 & 0.980 & 0.980 \\
    oh10.wc & 0.958 & 0.959 & 0.958 & 0.959 & 0.958 \\
    oh15.wc & 0.969 & 0.969 & 0.969 & 0.969 & 0.969 \\
    re0.wc & 0.920 & 0.919 & 0.917 & 0.917 & 0.917 \\
    re1.wc & 0.969 & 0.969 & 0.969 & 0.969 & 0.969 \\
    \midrule
    Average & 0.758 & 0.759 & 0.758 & 0.760 & 0.759 \\
    \bottomrule
    \end{tabular}%
  \label{tab:AW-Trees-timeouts2}%
\end{table}%


\begin{table}[htbp] \scriptsize
  \centering
  \caption{Results of predictive Specificity values of the baseline methods for inducing rule-based models.}
    \begin{tabular}{lrrrrr}
    \toprule
    data set & Decision Table & JRip  & OneR  & PART  & ZeroR \\
    \midrule
    CE    & 0.537 & 0.565 & 0.571 & 0.550 & 0.398 \\
    DM    & 0.416 & 0.456 & 0.382 & 0.511 & 0.327 \\
    MM    & 0.480 & 0.430 & 0.494 & 0.534 & 0.292 \\
    SC    & 0.345 & 0.378 & 0.221 & 0.458 & 0.161 \\
    DNA3  & 0.506 & 0.606 & 0.513 & 0.641 & 0.237 \\
    DNA11 & 0.284 & 0.373 & 0.370 & 0.565 & 0.244 \\
    PS    & 0.761 & 0.852 & 0.700 & 0.884 & 0.060 \\
    chen-2002 & 0.854 & 0.790 & 0.778 & 0.858 & 0.419 \\
    chowdary-2006 & 0.935 & 0.954 & 0.965 & 0.936 & 0.404 \\
    nutt-2003-v2 & 0.767 & 0.833 & 0.900 & 0.933 & 0.633 \\
    singh-2002 & 0.766 & 0.819 & 0.725 & 0.724 & 0.491 \\
    west-2001 & 0.848 & 0.935 & 0.918 & 0.815 & 0.490 \\
    dbworld-bodies & 0.786 & 0.743 & 0.865 & 0.743 & 0.455 \\
    dbworld-bodies-stemmed & 0.679 & 0.787 & 0.737 & 0.792 & 0.455 \\
    oh0.wc & 0.960 & 0.962 & 0.857 & 0.972 & 0.807 \\
    oh5.wc & 0.976 & 0.965 & 0.861 & 0.978 & 0.838 \\
    oh10.wc & 0.949 & 0.957 & 0.863 & 0.961 & 0.848 \\
    oh15.wc & 0.950 & 0.960 & 0.866 & 0.964 & 0.828 \\
    re0.wc & 0.852 & 0.923 & 0.712 & 0.922 & 0.596 \\
    re1.wc & 0.953 & 0.972 & 0.829 & 0.972 & 0.776 \\
    \midrule
    Average & 0.730 & 0.763 & 0.706 & 0.786 & 0.488 \\
    \bottomrule
    \end{tabular}%
\end{table}%

\begin{table}[htbp] \scriptsize
  \centering
  \caption{Results of predictive Sensitivity values of the baseline methods for inducing rule-based models.}
    \begin{tabular}{lrrrrr}
    \toprule
    data set & Decision Table & JRip  & OneR  & PART  & ZeroR \\
    \midrule
    CE    & 0.651 & 0.642 & 0.657 & 0.592 & 0.602 \\
    DM    & 0.647 & 0.673 & 0.655 & 0.639 & 0.673 \\
    MM    & 0.708 & 0.675 & 0.775 & 0.661 & 0.708 \\
    SC    & 0.851 & 0.811 & 0.835 & 0.806 & 0.839 \\
    DNA3  & 0.813 & 0.849 & 0.835 & 0.784 & 0.763 \\
    DNA11 & 0.732 & 0.718 & 0.732 & 0.755 & 0.756 \\
    PS    & 0.984 & 0.986 & 0.981 & 0.988 & 0.940 \\
    chen-2002 & 0.871 & 0.793 & 0.765 & 0.860 & 0.581 \\
    chowdary-2006 & 0.951 & 0.962 & 0.971 & 0.933 & 0.596 \\
    nutt-2003-v2 & 0.633 & 0.767 & 0.800 & 0.867 & 0.367 \\
    singh-2002 & 0.764 & 0.815 & 0.725 & 0.723 & 0.509 \\
    west-2001 & 0.835 & 0.915 & 0.915 & 0.835 & 0.510 \\
    dbworld-bodies & 0.798 & 0.740 & 0.860 & 0.740 & 0.545 \\
    dbworld-bodies-stemmed & 0.688 & 0.771 & 0.721 & 0.783 & 0.545 \\
    oh0.wc & 0.773 & 0.799 & 0.351 & 0.815 & 0.193 \\
    oh5.wc & 0.788 & 0.771 & 0.260 & 0.810 & 0.162 \\
    oh10.wc & 0.687 & 0.724 & 0.266 & 0.712 & 0.152 \\
    oh15.wc & 0.700 & 0.762 & 0.315 & 0.748 & 0.172 \\
    re0.wc & 0.668 & 0.781 & 0.543 & 0.725 & 0.404 \\
    re1.wc & 0.767 & 0.801 & 0.396 & 0.797 & 0.224 \\
    \midrule
    Average & 0.765 & 0.788 & 0.668 & 0.779 & 0.512 \\
    \bottomrule
    \end{tabular}%
\end{table}%

\begin{table}[htbp] \scriptsize
  \centering
  \caption{Results of predictive GMean values of the baseline methods for inducing rule-based models.}
    \begin{tabular}{lrrrrr}
    \toprule
    data set & Decision Table & JRip  & OneR  & PART  & ZeroR \\
    \midrule
    CE    & 0.591 & 0.602 & 0.612 & 0.571 & 0.489 \\
    DM    & 0.514 & 0.551 & 0.497 & 0.567 & 0.469 \\
    MM    & 0.567 & 0.531 & 0.610 & 0.589 & 0.451 \\
    SC    & 0.520 & 0.537 & 0.422 & 0.602 & 0.368 \\
    DNA3  & 0.629 & 0.715 & 0.653 & 0.707 & 0.424 \\
    DNA11 & 0.452 & 0.504 & 0.513 & 0.647 & 0.428 \\
    PS    & 0.865 & 0.916 & 0.828 & 0.935 & 0.238 \\
    chen-2002 & 0.862 & 0.791 & 0.771 & 0.859 & 0.493 \\
    chowdary-2006 & 0.943 & 0.958 & 0.968 & 0.934 & 0.490 \\
    nutt-2003-v2 & 0.692 & 0.794 & 0.845 & 0.898 & 0.477 \\
    singh-2002 & 0.765 & 0.817 & 0.725 & 0.723 & 0.500 \\
    west-2001 & 0.841 & 0.925 & 0.916 & 0.823 & 0.491 \\
    dbworld-bodies & 0.791 & 0.741 & 0.862 & 0.741 & 0.495 \\
    dbworld-bodies-stemmed & 0.683 & 0.779 & 0.729 & 0.787 & 0.495 \\
    oh0.wc & 0.861 & 0.876 & 0.548 & 0.890 & 0.395 \\
    oh5.wc & 0.877 & 0.862 & 0.473 & 0.890 & 0.369 \\
    oh10.wc & 0.807 & 0.832 & 0.479 & 0.827 & 0.359 \\
    oh15.wc & 0.815 & 0.855 & 0.522 & 0.849 & 0.377 \\
    re0.wc & 0.754 & 0.849 & 0.621 & 0.817 & 0.491 \\
    re1.wc & 0.855 & 0.882 & 0.573 & 0.880 & 0.417 \\
    \midrule
    average & 0.734 & 0.766 & 0.658 & 0.777 & 0.436 \\
    \bottomrule
    \end{tabular}%
\end{table}%


\begin{table}[htbp] \scriptsize
  \centering
  \caption{Results of predictive Specificity values of GGP-RI for timeouts from 1,000s to 5,000s.}
    \begin{tabular}{lrrrrr}
    \toprule
    data set & \multicolumn{1}{c}{1,000s} & \multicolumn{1}{c}{2,000s} & \multicolumn{1}{c}{3,000s} & \multicolumn{1}{c}{4,000s} & \multicolumn{1}{c}{5,000s} \\
    \midrule
    CE    & 0.413 & 0.415 & 0.415 & 0.415 & 0.415 \\
    DM    & 0.405 & 0.407 & 0.423 & 0.431 & 0.431 \\
    MM    & 0.399 & 0.411 & 0.423 & 0.423 & 0.426 \\
    SC    & 0.194 & 0.190 & 0.190 & 0.190 & 0.190 \\
    DNA3  & 0.429 & 0.428 & 0.427 & 0.427 & 0.427 \\
    DNA11 & 0.352 & 0.343 & 0.353 & 0.353 & 0.353 \\
    PS    & 0.280 & 0.297 & 0.284 & 0.268 & 0.281 \\
    chen-2002 & 0.632 & 0.628 & 0.617 & 0.624 & 0.624 \\
    chowdary-2006 & 0.809 & 0.814 & 0.814 & 0.811 & 0.808 \\
    nutt-2003-v2 & 0.670 & 0.670 & 0.660 & 0.660 & 0.660 \\
    singh-2002 & 0.618 & 0.619 & 0.621 & 0.626 & 0.634 \\
    west-2001 & 0.649 & 0.635 & 0.638 & 0.634 & 0.645 \\
    dbworld-bodies & 0.555 & 0.555 & 0.555 & 0.555 & 0.551 \\
    dbworld-bodies-stemmed & 0.640 & 0.640 & 0.644 & 0.644 & 0.644 \\
    oh0.wc & 0.809 & 0.809 & 0.809 & 0.809 & 0.809 \\
    oh5.wc & 0.846 & 0.845 & 0.845 & 0.845 & 0.845 \\
    oh10.wc & 0.843 & 0.847 & 0.847 & 0.847 & 0.847 \\
    oh15.wc & 0.833 & 0.833 & 0.833 & 0.833 & 0.833 \\
    re0.wc & 0.614 & 0.614 & 0.614 & 0.614 & 0.614 \\
    re1.wc & 0.786 & 0.786 & 0.786 & 0.787 & 0.787 \\
    \midrule
    Average & 0.589 & 0.589 & 0.590 & 0.590 & 0.591 \\
    \bottomrule
    \end{tabular}%
\end{table}%

\begin{table}[htbp] \scriptsize
  \centering
  \caption{Results of predictive Specificity values of GGP-RI for timeouts from 6,000s to 10,000s.}
    \begin{tabular}{lrrrrr}
    \toprule
    data set & \multicolumn{1}{c}{6,000s} & \multicolumn{1}{c}{7,000s} & \multicolumn{1}{c}{8,000s} & \multicolumn{1}{c}{9,000s} & \multicolumn{1}{c}{10,000s} \\
    \midrule
    CE    & 0.415 & 0.415 & 0.415 & 0.416 & 0.415 \\
    DM    & 0.431 & 0.430 & 0.430 & 0.429 & 0.429 \\
    MM    & 0.426 & 0.426 & 0.426 & 0.434 & 0.436 \\
    SC    & 0.187 & 0.187 & 0.187 & 0.187 & 0.190 \\
    DNA3  & 0.427 & 0.427 & 0.432 & 0.432 & 0.432 \\
    DNA11 & 0.353 & 0.353 & 0.353 & 0.363 & 0.363 \\
    PS    & 0.281 & 0.281 & 0.281 & 0.281 & 0.281 \\
    chen-2002 & 0.626 & 0.629 & 0.630 & 0.636 & 0.633 \\
    chowdary-2006 & 0.809 & 0.813 & 0.813 & 0.813 & 0.813 \\
    nutt-2003-v2 & 0.647 & 0.647 & 0.643 & 0.650 & 0.650 \\
    singh-2002 & 0.642 & 0.642 & 0.642 & 0.642 & 0.642 \\
    west-2001 & 0.649 & 0.649 & 0.649 & 0.649 & 0.647 \\
    dbworld-bodies & 0.551 & 0.551 & 0.551 & 0.551 & 0.558 \\
    dbworld-bodies-stemmed & 0.644 & 0.637 & 0.637 & 0.637 & 0.637 \\
    oh0.wc & 0.809 & 0.809 & 0.809 & 0.809 & 0.809 \\
    oh5.wc & 0.845 & 0.845 & 0.845 & 0.845 & 0.845 \\
    oh10.wc & 0.848 & 0.848 & 0.848 & 0.848 & 0.848 \\
    oh15.wc & 0.833 & 0.833 & 0.833 & 0.833 & 0.833 \\
    re0.wc & 0.614 & 0.613 & 0.613 & 0.614 & 0.614 \\
    re1.wc & 0.787 & 0.787 & 0.787 & 0.787 & 0.787 \\
    \midrule
    Average & 0.591 & 0.591 & 0.591 & 0.593 & 0.593 \\
    \bottomrule
    \end{tabular}%
\end{table}%

\begin{table}[htbp] \scriptsize
  \centering
  \caption{Results of predictive Sensitivity values of GGP-RI for timeouts from 1,000s to 5,000s.}
    \begin{tabular}{lrrrrr}
    \toprule
    data set & \multicolumn{1}{c}{1,000s} & \multicolumn{1}{c}{2,000s} & \multicolumn{1}{c}{3,000s} & \multicolumn{1}{c}{4,000s} & \multicolumn{1}{c}{5,000s} \\
    \midrule
    CE    & 0.607 & 0.608 & 0.608 & 0.609 & 0.608 \\
    DM    & 0.684 & 0.684 & 0.693 & 0.698 & 0.698 \\
    MM    & 0.717 & 0.722 & 0.726 & 0.726 & 0.726 \\
    SC    & 0.820 & 0.818 & 0.818 & 0.821 & 0.821 \\
    DNA3  & 0.813 & 0.812 & 0.807 & 0.807 & 0.807 \\
    DNA11 & 0.739 & 0.738 & 0.741 & 0.741 & 0.741 \\
    PS    & 0.952 & 0.953 & 0.952 & 0.951 & 0.952 \\
    chen-2002 & 0.686 & 0.684 & 0.676 & 0.683 & 0.683 \\
    chowdary-2006 & 0.854 & 0.858 & 0.858 & 0.855 & 0.854 \\
    nutt-2003-v2 & 0.620 & 0.620 & 0.600 & 0.600 & 0.600 \\
    singh-2002 & 0.609 & 0.613 & 0.615 & 0.619 & 0.626 \\
    west-2001 & 0.594 & 0.585 & 0.589 & 0.593 & 0.602 \\
    dbworld-bodies & 0.615 & 0.615 & 0.615 & 0.615 & 0.612 \\
    dbworld-bodies-stemmed & 0.669 & 0.669 & 0.671 & 0.671 & 0.671 \\
    oh0.wc & 0.196 & 0.196 & 0.196 & 0.196 & 0.196 \\
    oh5.wc & 0.156 & 0.158 & 0.158 & 0.158 & 0.158 \\
    oh10.wc & 0.162 & 0.160 & 0.160 & 0.160 & 0.160 \\
    oh15.wc & 0.175 & 0.175 & 0.175 & 0.175 & 0.174 \\
    re0.wc & 0.399 & 0.399 & 0.399 & 0.399 & 0.399 \\
    re1.wc & 0.217 & 0.218 & 0.218 & 0.218 & 0.217 \\
    \midrule
    Average & 0.564 & 0.564 & 0.564 & 0.565 & 0.565 \\
    \bottomrule
    \end{tabular}%
\end{table}%

\begin{table}[htbp] \scriptsize
  \centering
  \caption{Results of predictive Sensitivity values of GGP-RI for timeouts from 6,000s to 10,000s.}
    \begin{tabular}{lrrrrr}
    \toprule
    data set & \multicolumn{1}{c}{6,000s} & \multicolumn{1}{c}{7,000s} & \multicolumn{1}{c}{8,000s} & \multicolumn{1}{c}{9,000s} & \multicolumn{1}{c}{10,000s} \\
    \midrule
    CE    & 0.608 & 0.608 & 0.608 & 0.608 & 0.608 \\
    DM    & 0.698 & 0.702 & 0.702 & 0.700 & 0.700 \\
    MM    & 0.726 & 0.726 & 0.726 & 0.726 & 0.724 \\
    SC    & 0.823 & 0.823 & 0.823 & 0.823 & 0.821 \\
    DNA3  & 0.807 & 0.807 & 0.809 & 0.809 & 0.809 \\
    DNA11 & 0.741 & 0.741 & 0.741 & 0.741 & 0.741 \\
    PS    & 0.952 & 0.952 & 0.952 & 0.952 & 0.952 \\
    chen-2002 & 0.684 & 0.685 & 0.686 & 0.692 & 0.689 \\
    chowdary-2006 & 0.855 & 0.857 & 0.857 & 0.857 & 0.857 \\
    nutt-2003-v2 & 0.593 & 0.593 & 0.587 & 0.600 & 0.600 \\
    singh-2002 & 0.634 & 0.634 & 0.634 & 0.634 & 0.634 \\
    west-2001 & 0.614 & 0.614 & 0.614 & 0.614 & 0.610 \\
    dbworld-bodies & 0.612 & 0.612 & 0.612 & 0.612 & 0.619 \\
    dbworld-bodies-stemmed & 0.671 & 0.665 & 0.665 & 0.665 & 0.665 \\
    oh0.wc & 0.196 & 0.196 & 0.196 & 0.196 & 0.196 \\
    oh5.wc & 0.158 & 0.158 & 0.158 & 0.158 & 0.158 \\
    oh10.wc & 0.162 & 0.162 & 0.162 & 0.162 & 0.162 \\
    oh15.wc & 0.175 & 0.175 & 0.175 & 0.175 & 0.175 \\
    re0.wc & 0.399 & 0.398 & 0.398 & 0.399 & 0.399 \\
    re1.wc & 0.217 & 0.217 & 0.217 & 0.217 & 0.217 \\
    \midrule
    Average & 0.566 & 0.566 & 0.566 & 0.567 & 0.567 \\
    \bottomrule
    \end{tabular}%
\end{table}%

\begin{table}[htbp] \scriptsize
  \centering
  \caption{Results of predictive Gmean values of GGP-RI for timeouts from 1,000s to 5,000s.}
    \begin{tabular}{lrrrrr}
    \toprule
    data set & \multicolumn{1}{c}{1,000s} & \multicolumn{1}{c}{2,000s} & \multicolumn{1}{c}{3,000s} & \multicolumn{1}{c}{4,000s} & \multicolumn{1}{c}{5,000s} \\
    \midrule
    CE    & 0.501 & 0.502 & 0.502 & 0.503 & 0.503 \\
    DM    & 0.523 & 0.525 & 0.538 & 0.545 & 0.545 \\
    MM    & 0.524 & 0.532 & 0.542 & 0.542 & 0.544 \\
    SC    & 0.392 & 0.388 & 0.388 & 0.389 & 0.389 \\
    DNA3  & 0.582 & 0.581 & 0.578 & 0.578 & 0.578 \\
    DNA11 & 0.498 & 0.491 & 0.499 & 0.499 & 0.499 \\
    PS    & 0.445 & 0.463 & 0.451 & 0.436 & 0.447 \\
    chen-2002 & 0.658 & 0.654 & 0.645 & 0.652 & 0.652 \\
    chowdary-2006 & 0.83  & 0.834 & 0.834 & 0.832 & 0.829 \\
    nutt-2003-v2 & 0.631 & 0.631 & 0.616 & 0.616 & 0.616 \\
    singh-2002 & 0.613 & 0.616 & 0.618 & 0.623 & 0.63 \\
    west-2001 & 0.617 & 0.605 & 0.609 & 0.609 & 0.619 \\
    dbworld-bodies & 0.582 & 0.582 & 0.582 & 0.582 & 0.578 \\
    dbworld-bodies-stemmed & 0.652 & 0.652 & 0.655 & 0.655 & 0.655 \\
    oh0.wc & 0.398 & 0.398 & 0.398 & 0.398 & 0.398 \\
    oh5.wc & 0.361 & 0.364 & 0.364 & 0.364 & 0.364 \\
    oh10.wc & 0.37  & 0.367 & 0.367 & 0.367 & 0.367 \\
    oh15.wc & 0.382 & 0.381 & 0.381 & 0.381 & 0.381 \\
    re0.wc & 0.489 & 0.489 & 0.489 & 0.489 & 0.489 \\
    re1.wc & 0.407 & 0.408 & 0.408 & 0.408 & 0.407 \\
    \midrule
    Average & 0.523 & 0.523 & 0.523 & 0.523 & 0.525 \\
    \bottomrule
    \end{tabular}%
\end{table}%

\begin{table}[htbp] \scriptsize
  \centering
  \caption{Results of predictive GMean values of GGP-RI for timeouts from 6,000s to 10,000s.}
    \begin{tabular}{lrrrrr}
    \toprule
    data set & \multicolumn{1}{c}{6,000s} & \multicolumn{1}{c}{7,000s} & \multicolumn{1}{c}{8,000s} & \multicolumn{1}{c}{9,000s} & \multicolumn{1}{c}{10,000s} \\
    \midrule
    CE    & 0.503 & 0.503 & 0.503 & 0.503 & 0.502 \\
    DM    & 0.545 & 0.546 & 0.546 & 0.544 & 0.544 \\
    MM    & 0.544 & 0.544 & 0.544 & 0.549 & 0.55 \\
    SC    & 0.386 & 0.386 & 0.386 & 0.386 & 0.389 \\
    DNA3  & 0.578 & 0.578 & 0.583 & 0.583 & 0.583 \\
    DNA11 & 0.499 & 0.499 & 0.499 & 0.506 & 0.506 \\
    PS    & 0.448 & 0.448 & 0.448 & 0.448 & 0.448 \\
    chen-2002 & 0.653 & 0.655 & 0.657 & 0.663 & 0.659 \\
    chowdary-2006 & 0.831 & 0.833 & 0.833 & 0.833 & 0.833 \\
    nutt-2003-v2 & 0.606 & 0.606 & 0.6   & 0.611 & 0.611 \\
    singh-2002 & 0.638 & 0.638 & 0.638 & 0.638 & 0.638 \\
    west-2001 & 0.628 & 0.628 & 0.628 & 0.628 & 0.624 \\
    dbworld-bodies & 0.578 & 0.578 & 0.578 & 0.578 & 0.585 \\
    dbworld-bodies-stemmed & 0.655 & 0.649 & 0.649 & 0.649 & 0.649 \\
    oh0.wc & 0.398 & 0.398 & 0.398 & 0.398 & 0.398 \\
    oh5.wc & 0.364 & 0.364 & 0.364 & 0.364 & 0.364 \\
    oh10.wc & 0.369 & 0.369 & 0.369 & 0.369 & 0.369 \\
    oh15.wc & 0.381 & 0.381 & 0.381 & 0.381 & 0.381 \\
    re0.wc & 0.489 & 0.489 & 0.489 & 0.489 & 0.489 \\
    re1.wc & 0.407 & 0.407 & 0.407 & 0.407 & 0.407 \\
    \midrule
    Average & 0.525 & 0.525 & 0.525 & 0.526 & 0.526 \\
    \bottomrule
    \end{tabular}%
\end{table}%

\begin{table}[htbp] \scriptsize
  \centering
  \caption{Results of predictive Specificity values of Auto-WEKA-Rules for timeouts from 1,000s to 5,000s.}
    \begin{tabular}{lrrrrr}
    \toprule
    data set & \multicolumn{1}{c}{1,000s} & \multicolumn{1}{c}{2,000s} & \multicolumn{1}{c}{3,000s} & \multicolumn{1}{c}{4,000s} & \multicolumn{1}{c}{5,000s} \\
    \midrule
    CE    & 0.532 & 0.529 & 0.531 & 0.532 & 0.532 \\
    DM    & 0.360 & 0.361 & 0.368 & 0.358 & 0.362 \\
    MM    & 0.319 & 0.317 & 0.316 & 0.316 & 0.316 \\
    SC    & 0.223 & 0.233 & 0.233 & 0.228 & 0.233 \\
    DNA3  & 0.532 & 0.555 & 0.562 & 0.588 & 0.571 \\
    DNA11 & 0.382 & 0.384 & 0.400 & 0.435 & 0.399 \\
    PS    & 0.676 & 0.697 & 0.697 & 0.709 & 0.706 \\
    chen-2002 & 0.850 & 0.867 & 0.850 & 0.853 & 0.850 \\
    chowdary-2006 & 0.938 & 0.930 & 0.928 & 0.928 & 0.928 \\
    nutt-2003-v2 & 0.804 & 0.808 & 0.792 & 0.792 & 0.792 \\
    singh-2002 & 0.791 & 0.792 & 0.798 & 0.791 & 0.808 \\
    west-2001 & 0.927 & 0.927 & 0.927 & 0.927 & 0.927 \\
    dbworld-bodies & 0.783 & 0.763 & 0.777 & 0.793 & 0.783 \\
    dbworld-bodies-stemmed & 0.728 & 0.759 & 0.771 & 0.764 & 0.781 \\
    oh0.wc & 0.949 & 0.952 & 0.955 & 0.956 & 0.956 \\
    oh5.wc & 0.967 & 0.971 & 0.973 & 0.975 & 0.975 \\
    oh10.wc & 0.952 & 0.953 & 0.952 & 0.952 & 0.951 \\
    oh15.wc & 0.955 & 0.957 & 0.957 & 0.957 & 0.958 \\
    re0.wc & 0.851 & 0.911 & 0.906 & 0.906 & 0.907 \\
    re1.wc & 0.912 & 0.945 & 0.951 & 0.950 & 0.950 \\
    \midrule
    Average & 0.722 & 0.731 & 0.732 & 0.735 & 0.734 \\
    \bottomrule
    \end{tabular}%
\end{table}%

\begin{table}[htbp] \scriptsize
  \centering
  \caption{Results of predictive Specificity values of Auto-WEKA-Rules for timeouts from 6,000s to 10,000s.}
    \begin{tabular}{lrrrrr}
    \toprule
    data set & \multicolumn{1}{c}{6,000s} & \multicolumn{1}{c}{7,000s} & \multicolumn{1}{c}{8,000s} & \multicolumn{1}{c}{9,000s} & \multicolumn{1}{c}{10,000s} \\
    \midrule
    CE    & 0.532 & 0.537 & 0.534 & 0.534 & 0.535 \\
    DM    & 0.365 & 0.379 & 0.371 & 0.376 & 0.363 \\
    MM    & 0.316 & 0.316 & 0.316 & 0.316 & 0.316 \\
    SC    & 0.238 & 0.233 & 0.228 & 0.253 & 0.212 \\
    DNA3  & 0.571 & 0.554 & 0.564 & 0.568 & 0.549 \\
    DNA11 & 0.439 & 0.452 & 0.427 & 0.421 & 0.448 \\
    PS    & 0.712 & 0.712 & 0.720 & 0.730 & 0.729 \\
    chen-2002 & 0.866 & 0.859 & 0.853 & 0.864 & 0.857 \\
    chowdary-2006 & 0.928 & 0.928 & 0.928 & 0.928 & 0.928 \\
    nutt-2003-v2 & 0.808 & 0.792 & 0.792 & 0.804 & 0.808 \\
    singh-2002 & 0.798 & 0.805 & 0.808 & 0.796 & 0.800 \\
    west-2001 & 0.919 & 0.919 & 0.919 & 0.919 & 0.919 \\
    dbworld-bodies & 0.776 & 0.782 & 0.780 & 0.782 & 0.782 \\
    dbworld-bodies-stemmed & 0.797 & 0.770 & 0.774 & 0.788 & 0.794 \\
    oh0.wc & 0.957 & 0.956 & 0.955 & 0.956 & 0.956 \\
    oh5.wc & 0.977 & 0.975 & 0.976 & 0.977 & 0.977 \\
    oh10.wc & 0.952 & 0.953 & 0.953 & 0.953 & 0.955 \\
    oh15.wc & 0.957 & 0.957 & 0.957 & 0.957 & 0.957 \\
    re0.wc & 0.909 & 0.908 & 0.908 & 0.906 & 0.908 \\
    re1.wc & 0.950 & 0.951 & 0.950 & 0.951 & 0.951 \\
    \midrule
    Average & 0.738 & 0.737 & 0.736 & 0.739 & 0.737 \\
    \bottomrule
    \end{tabular}%
\end{table}%

\begin{table}[htbp] \scriptsize
  \centering
  \caption{Results of predictive Sensitivity values of Auto-WEKA-Rules for timeouts from 1,000s to 5,000s.}
    \begin{tabular}{lrrrrr}
    \toprule
    data set & \multicolumn{1}{c}{1,000s} & \multicolumn{1}{c}{2,000s} & \multicolumn{1}{c}{3,000s} & \multicolumn{1}{c}{4,000s} & \multicolumn{1}{c}{5,000s} \\
    \midrule
    CE    & 0.628 & 0.626 & 0.629 & 0.629 & 0.631 \\
    DM    & 0.649 & 0.651 & 0.653 & 0.651 & 0.653 \\
    MM    & 0.694 & 0.692 & 0.689 & 0.689 & 0.689 \\
    SC    & 0.820 & 0.818 & 0.816 & 0.815 & 0.816 \\
    DNA3  & 0.835 & 0.840 & 0.840 & 0.844 & 0.840 \\
    DNA11 & 0.719 & 0.730 & 0.725 & 0.741 & 0.734 \\
    PS    & 0.977 & 0.977 & 0.977 & 0.977 & 0.978 \\
    chen-2002 & 0.871 & 0.889 & 0.875 & 0.875 & 0.871 \\
    chowdary-2006 & 0.945 & 0.941 & 0.938 & 0.938 & 0.938 \\
    nutt-2003-v2 & 0.783 & 0.792 & 0.783 & 0.783 & 0.783 \\
    singh-2002 & 0.793 & 0.793 & 0.798 & 0.791 & 0.808 \\
    west-2001 & 0.915 & 0.915 & 0.915 & 0.915 & 0.915 \\
    dbworld-bodies & 0.792 & 0.772 & 0.783 & 0.796 & 0.788 \\
    dbworld-bodies-stemmed & 0.729 & 0.758 & 0.758 & 0.759 & 0.767 \\
    oh0.wc & 0.750 & 0.762 & 0.768 & 0.772 & 0.771 \\
    oh5.wc & 0.751 & 0.771 & 0.774 & 0.780 & 0.781 \\
    oh10.wc & 0.702 & 0.704 & 0.705 & 0.702 & 0.701 \\
    oh15.wc & 0.751 & 0.753 & 0.754 & 0.755 & 0.756 \\
    re0.wc & 0.671 & 0.744 & 0.740 & 0.740 & 0.740 \\
    re1.wc & 0.610 & 0.720 & 0.755 & 0.754 & 0.754 \\
    \midrule
    Average & 0.769 & 0.782 & 0.784 & 0.785 & 0.786 \\
    \bottomrule
    \end{tabular}%
\end{table}%

\begin{table}[htbp] \scriptsize
  \centering
  \caption{Results of predictive Sensitivity values of Auto-WEKA-Rules for timeouts from 6,000s to 10,000s.}
    \begin{tabular}{lrrrrr}
    \toprule
    data set & \multicolumn{1}{c}{6,000s} & \multicolumn{1}{c}{7,000s} & \multicolumn{1}{c}{8,000s} & \multicolumn{1}{c}{9,000s} & \multicolumn{1}{c}{10,000s} \\
    \midrule
    CE    & 0.630 & 0.634 & 0.633 & 0.633 & 0.633 \\
    DM    & 0.660 & 0.662 & 0.657 & 0.649 & 0.649 \\
    MM    & 0.689 & 0.689 & 0.689 & 0.689 & 0.689 \\
    SC    & 0.819 & 0.815 & 0.815 & 0.815 & 0.815 \\
    DNA3  & 0.840 & 0.835 & 0.831 & 0.835 & 0.831 \\
    DNA11 & 0.751 & 0.747 & 0.736 & 0.746 & 0.744 \\
    PS    & 0.978 & 0.978 & 0.978 & 0.979 & 0.979 \\
    chen-2002 & 0.883 & 0.879 & 0.872 & 0.881 & 0.878 \\
    chowdary-2006 & 0.938 & 0.938 & 0.938 & 0.938 & 0.938 \\
    nutt-2003-v2 & 0.792 & 0.783 & 0.783 & 0.783 & 0.792 \\
    singh-2002 & 0.798 & 0.805 & 0.808 & 0.795 & 0.801 \\
    west-2001 & 0.910 & 0.910 & 0.910 & 0.910 & 0.910 \\
    dbworld-bodies & 0.784 & 0.789 & 0.789 & 0.789 & 0.789 \\
    dbworld-bodies-stemmed & 0.784 & 0.759 & 0.763 & 0.775 & 0.783 \\
    oh0.wc & 0.774 & 0.772 & 0.771 & 0.775 & 0.773 \\
    oh5.wc & 0.786 & 0.780 & 0.782 & 0.786 & 0.786 \\
    oh10.wc & 0.704 & 0.706 & 0.707 & 0.709 & 0.712 \\
    oh15.wc & 0.757 & 0.757 & 0.755 & 0.756 & 0.757 \\
    re0.wc & 0.744 & 0.743 & 0.745 & 0.740 & 0.743 \\
    re1.wc & 0.754 & 0.755 & 0.754 & 0.755 & 0.755 \\
    \midrule
    Average & 0.789 & 0.787 & 0.786 & 0.787 & 0.788 \\
    \bottomrule
    \end{tabular}%
\end{table}%

\begin{table}[htbp] \scriptsize
  \centering
  \caption{Results of predictive GMean values of Auto-WEKA-Rules for timeouts from 1,000s to 5,000s.}
    \begin{tabular}{lrrrrr}
    \toprule
    data set & \multicolumn{1}{c}{1,000s} & \multicolumn{1}{c}{2,000s} & \multicolumn{1}{c}{3,000s} & \multicolumn{1}{c}{4,000s} & \multicolumn{1}{c}{5,000s} \\
    \midrule
    CE    & 0.578 & 0.575 & 0.578 & 0.578 & 0.579 \\
    DM    & 0.480 & 0.482 & 0.487 & 0.480 & 0.483 \\
    MM    & 0.465 & 0.463 & 0.462 & 0.462 & 0.462 \\
    SC    & 0.419 & 0.427 & 0.426 & 0.421 & 0.426 \\
    DNA3  & 0.658 & 0.674 & 0.678 & 0.695 & 0.683 \\
    DNA11 & 0.517 & 0.521 & 0.528 & 0.558 & 0.533 \\
    PS    & 0.793 & 0.805 & 0.805 & 0.811 & 0.810 \\
    chen-2002 & 0.860 & 0.878 & 0.862 & 0.864 & 0.860 \\
    chowdary-2006 & 0.941 & 0.935 & 0.933 & 0.933 & 0.933 \\
    nutt-2003-v2 & 0.789 & 0.796 & 0.783 & 0.783 & 0.783 \\
    singh-2002 & 0.792 & 0.793 & 0.798 & 0.791 & 0.808 \\
    west-2001 & 0.920 & 0.920 & 0.920 & 0.920 & 0.920 \\
    dbworld-bodies & 0.787 & 0.767 & 0.780 & 0.795 & 0.785 \\
    dbworld-bodies-stemmed & 0.728 & 0.758 & 0.764 & 0.761 & 0.774 \\
    oh0.wc & 0.843 & 0.852 & 0.856 & 0.858 & 0.858 \\
    oh5.wc & 0.852 & 0.865 & 0.867 & 0.872 & 0.872 \\
    oh10.wc & 0.817 & 0.819 & 0.819 & 0.817 & 0.817 \\
    oh15.wc & 0.847 & 0.849 & 0.849 & 0.850 & 0.850 \\
    re0.wc & 0.756 & 0.823 & 0.819 & 0.819 & 0.819 \\
    re1.wc & 0.742 & 0.823 & 0.847 & 0.846 & 0.846 \\
    \midrule
    Average & 0.729 & 0.741 & 0.743 & 0.746 & 0.745 \\
    \bottomrule
    \end{tabular}%
\end{table}%

\begin{table}[htbp] \scriptsize
  \centering
  \caption{Results of predictive GMean values of Auto-WEKA-Rules for timeouts from 6,000s to 10,000s.}
    \begin{tabular}{lrrrrr}
    \toprule
    data set & \multicolumn{1}{c}{6,000s} & \multicolumn{1}{c}{7,000s} & \multicolumn{1}{c}{8,000s} & \multicolumn{1}{c}{9,000s} & \multicolumn{1}{c}{10,000s} \\
    \midrule
    CE    & 0.579 & 0.584 & 0.581 & 0.582 & 0.582 \\
    DM    & 0.487 & 0.497 & 0.490 & 0.489 & 0.482 \\
    MM    & 0.462 & 0.462 & 0.462 & 0.462 & 0.462 \\
    SC    & 0.432 & 0.426 & 0.421 & 0.443 & 0.406 \\
    DNA3  & 0.683 & 0.670 & 0.675 & 0.680 & 0.668 \\
    DNA11 & 0.567 & 0.572 & 0.550 & 0.551 & 0.567 \\
    PS    & 0.813 & 0.813 & 0.817 & 0.823 & 0.823 \\
    chen-2002 & 0.874 & 0.869 & 0.862 & 0.872 & 0.867 \\
    chowdary-2006 & 0.933 & 0.933 & 0.933 & 0.933 & 0.933 \\
    nutt-2003-v2 & 0.796 & 0.783 & 0.783 & 0.789 & 0.796 \\
    singh-2002 & 0.798 & 0.805 & 0.808 & 0.795 & 0.800 \\
    west-2001 & 0.914 & 0.914 & 0.914 & 0.914 & 0.914 \\
    dbworld-bodies & 0.780 & 0.785 & 0.784 & 0.785 & 0.785 \\
    dbworld-bodies-stemmed & 0.790 & 0.764 & 0.768 & 0.781 & 0.788 \\
    oh0.wc & 0.860 & 0.859 & 0.858 & 0.861 & 0.859 \\
    oh5.wc & 0.876 & 0.872 & 0.874 & 0.876 & 0.876 \\
    oh10.wc & 0.819 & 0.820 & 0.821 & 0.822 & 0.824 \\
    oh15.wc & 0.851 & 0.851 & 0.850 & 0.850 & 0.851 \\
    re0.wc & 0.822 & 0.822 & 0.822 & 0.819 & 0.821 \\
    re1.wc & 0.846 & 0.847 & 0.846 & 0.847 & 0.847 \\
    \midrule
    Average & 0.749 & 0.747 & 0.746 & 0.749 & 0.748 \\
    \bottomrule
    \end{tabular}%
\end{table}%


\begin{table}[htbp] \scriptsize
  \centering
  \caption{Results of predictive Specificity values of the baseline methods for building Bayesian network classifiers.}
    \begin{tabular}{lrrr}
    \toprule
    data set & BayesNet & NaiveBayes & \multicolumn{1}{c}{NaiveBayesMultinomial} \\
    \midrule
    CE    & 0.550 & 0.554 & 0.556 \\
    DM    & 0.545 & 0.566 & 0.620 \\
    MM    & 0.498 & 0.507 & 0.574 \\
    SC    & 0.525 & 0.468 & 0.609 \\
    DNA3  & 0.736 & 0.716 & 0.742 \\
    DNA11 & 0.672 & 0.672 & 0.509 \\
    PS    & 0.800 & 0.779 & 0.772 \\
    chen-2002 & 0.884 & 0.924 & 0.864 \\
    chowdary-2006 & 0.954 & 0.980 & 0.980 \\
    nutt-2003-v2 & 0.800 & 0.750 & 0.367 \\
    singh-2002 & 0.833 & 0.785 & 0.725 \\
    west-2001 & 0.892 & 0.875 & 0.835 \\
    dbworld-bodies & 0.811 & 0.700 & 0.905 \\
    dbworld-bodies-stemmed & 0.733 & 0.733 & 0.871 \\
    oh0.wc & 0.987 & 0.977 & 0.988 \\
    oh5.wc & 0.983 & 0.973 & 0.984 \\
    oh10.wc & 0.976 & 0.964 & 0.974 \\
    oh15.wc & 0.977 & 0.965 & 0.978 \\
    re0.wc & 0.930 & 0.938 & 0.949 \\
    re1.wc & 0.976 & 0.974 & 0.975 \\
    \midrule
    Average & 0.803 & 0.790 & 0.789 \\
    \bottomrule
    \end{tabular}%
\end{table}%

\begin{table}[htbp] \scriptsize
  \centering
  \caption{Results of predictive Sensitivity values of the baseline methods for building Bayesian network classifiers.}
    \begin{tabular}{lrrr}
    \toprule
    data set & BayesNet & NaiveBayes & \multicolumn{1}{c}{NaiveBayesMultinomial} \\
    \midrule
    CE    & 0.586 & 0.598 & 0.584 \\
    DM    & 0.647 & 0.664 & 0.647 \\
    MM    & 0.650 & 0.683 & 0.707 \\
    SC    & 0.738 & 0.754 & 0.750 \\
    DNA3  & 0.805 & 0.820 & 0.829 \\
    DNA11 & 0.734 & 0.734 & 0.689 \\
    PS    & 0.974 & 0.981 & 0.979 \\
    chen-2002 & 0.905 & 0.933 & 0.899 \\
    chowdary-2006 & 0.962 & 0.981 & 0.981 \\
    nutt-2003-v2 & 0.700 & 0.700 & 0.333 \\
    singh-2002 & 0.834 & 0.785 & 0.725 \\
    west-2001 & 0.875 & 0.875 & 0.815 \\
    dbworld-bodies & 0.814 & 0.750 & 0.895 \\
    dbworld-bodies-stemmed & 0.767 & 0.767 & 0.862 \\
    oh0.wc & 0.899 & 0.796 & 0.895 \\
    oh5.wc & 0.855 & 0.787 & 0.869 \\
    oh10.wc & 0.804 & 0.721 & 0.801 \\
    oh15.wc & 0.829 & 0.725 & 0.834 \\
    re0.wc & 0.759 & 0.566 & 0.798 \\
    re1.wc & 0.814 & 0.664 & 0.837 \\
    \midrule
    Average & 0.798 & 0.764 & 0.786 \\
    \bottomrule
    \end{tabular}%
\end{table}%

\begin{table}[htbp] \scriptsize
  \centering
  \caption{Results of predictive GMean values of the baseline methods for building Bayesian network classifiers.}
    \begin{tabular}{lrrr}
    \toprule
    data set & BayesNet & NaiveBayes & \multicolumn{1}{c}{NaiveBayesMultinomial} \\
    \midrule
    CE    & 0.567 & 0.576 & 0.569 \\
    DM    & 0.589 & 0.608 & 0.632 \\
    MM    & 0.560 & 0.578 & 0.627 \\
    SC    & 0.616 & 0.587 & 0.672 \\
    DNA3  & 0.766 & 0.762 & 0.778 \\
    DNA11 & 0.699 & 0.699 & 0.581 \\
    PS    & 0.883 & 0.874 & 0.869 \\
    chen-2002 & 0.894 & 0.928 & 0.881 \\
    chowdary-2006 & 0.958 & 0.980 & 0.980 \\
    nutt-2003-v2 & 0.745 & 0.716 & 0.340 \\
    singh-2002 & 0.833 & 0.785 & 0.725 \\
    west-2001 & 0.883 & 0.874 & 0.824 \\
    dbworld-bodies & 0.812 & 0.722 & 0.900 \\
    dbworld-bodies-stemmed & 0.749 & 0.749 & 0.866 \\
    oh0.wc & 0.942 & 0.882 & 0.940 \\
    oh5.wc & 0.917 & 0.875 & 0.925 \\
    oh10.wc & 0.886 & 0.833 & 0.883 \\
    oh15.wc & 0.900 & 0.836 & 0.903 \\
    re0.wc & 0.840 & 0.728 & 0.870 \\
    re1.wc & 0.891 & 0.804 & 0.903 \\
    \midrule
    Average & 0.797 & 0.770 & 0.784 \\
    \bottomrule
    \end{tabular}%
\end{table}%

\begin{table}[htbp] \scriptsize
  \centering
  \caption{Results of predictive Specificity values of HHEA-BNC for timeouts from 1,000s to 5,000s.}
    \begin{tabular}{lrrrrr}
    \toprule
    data set & \multicolumn{1}{c}{1,000s} & \multicolumn{1}{c}{2,000s} & \multicolumn{1}{c}{3,000s} & \multicolumn{1}{c}{4,000s} & \multicolumn{1}{c}{5,000s} \\
    \midrule
    CE    & 0.540 & 0.541 & 0.539 & 0.533 & 0.534 \\
    DM    & 0.508 & 0.553 & 0.571 & 0.552 & 0.544 \\
    MM    & 0.562 & 0.560 & 0.572 & 0.565 & 0.549 \\
    SC    & 0.332 & 0.328 & 0.343 & 0.348 & 0.356 \\
    DNA3 & 0.670 & 0.672 & 0.674 & 0.673 & 0.666 \\
    DNA11 & 0.423 & 0.389 & 0.384 & 0.380 & 0.369 \\
    PS    & 0.724 & 0.737 & 0.751 & 0.745 & 0.745 \\
    chen-2002 & 0.841 & 0.870 & 0.865 & 0.867 & 0.864 \\
    chowdary-2006 & 0.962 & 0.962 & 0.959 & 0.962 & 0.959 \\
    nutt-2003-v2 & 0.780 & 0.820 & 0.810 & 0.813 & 0.840 \\
    singh-2002 & 0.771 & 0.778 & 0.780 & 0.793 & 0.791 \\
    west-2001 & 0.885 & 0.884 & 0.878 & 0.878 & 0.894 \\
    dbworld-bodies & 0.742 & 0.742 & 0.759 & 0.780 & 0.775 \\
    dbworld-bodies-stemmed & 0.758 & 0.797 & 0.809 & 0.778 & 0.812 \\
    oh0.wc & 0.986 & 0.986 & 0.986 & 0.986 & 0.986 \\
    oh5.wc & 0.982 & 0.979 & 0.972 & 0.975 & 0.975 \\
    oh10.wc & 0.972 & 0.974 & 0.971 & 0.972 & 0.969 \\
    oh15.wc & 0.980 & 0.977 & 0.977 & 0.974 & 0.980 \\
    re0.wc & 0.931 & 0.931 & 0.930 & 0.930 & 0.931 \\
    re1.wc & 0.938 & 0.968 & 0.970 & 0.971 & 0.971 \\
    \midrule
    Average & 0.764 & 0.772 & 0.775 & 0.774 & 0.775 \\
    \bottomrule
    \end{tabular}%
\end{table}%

\begin{table}[htbp] \scriptsize
  \centering
  \caption{Results of predictive Specificity values of HHEA-BNC for timeouts from 6,000s to 10,000s.}
    \begin{tabular}{lrrrrr}
    \toprule
    data set & \multicolumn{1}{c}{6,000s} & \multicolumn{1}{c}{7,000s} & \multicolumn{1}{c}{8,000s} & \multicolumn{1}{c}{9,000s} & \multicolumn{1}{c}{10,000s} \\
    \midrule
    CE    & 0.541 & 0.538 & 0.539 & 0.544 & 0.544 \\
    DM    & 0.555 & 0.550 & 0.579 & 0.567 & 0.563 \\
    MM    & 0.525 & 0.520 & 0.512 & 0.505 & 0.510 \\
    SC    & 0.360 & 0.343 & 0.364 & 0.381 & 0.365 \\
    DNA3 & 0.670 & 0.667 & 0.650 & 0.650 & 0.649 \\
    DNA11 & 0.357 & 0.361 & 0.354 & 0.357 & 0.356 \\
    PS    & 0.747 & 0.739 & 0.740 & 0.748 & 0.729 \\
    chen-2002 & 0.863 & 0.865 & 0.862 & 0.862 & 0.858 \\
    chowdary-2006 & 0.962 & 0.965 & 0.965 & 0.962 & 0.951 \\
    nutt-2003-v2 & 0.827 & 0.843 & 0.847 & 0.837 & 0.840 \\
    singh-2002 & 0.789 & 0.785 & 0.783 & 0.786 & 0.776 \\
    west-2001 & 0.884 & 0.886 & 0.882 & 0.882 & 0.877 \\
    dbworld-bodies & 0.753 & 0.771 & 0.758 & 0.787 & 0.776 \\
    dbworld-bodies-stemmed & 0.803 & 0.812 & 0.791 & 0.801 & 0.798 \\
    oh0.wc & 0.986 & 0.983 & 0.979 & 0.979 & 0.979 \\
    oh5.wc & 0.977 & 0.972 & 0.972 & 0.972 & 0.978 \\
    oh10.wc & 0.966 & 0.966 & 0.963 & 0.966 & 0.966 \\
    oh15.wc & 0.980 & 0.980 & 0.980 & 0.980 & 0.977 \\
    re0.wc & 0.931 & 0.931 & 0.918 & 0.917 & 0.917 \\
    re1.wc & 0.971 & 0.971 & 0.972 & 0.972 & 0.972 \\
    \midrule
    Average & 0.772 & 0.772 & 0.770 & 0.773 & 0.769 \\
    \bottomrule
    \end{tabular}%
\end{table}%

\begin{table}[htbp] \scriptsize
  \centering
  \caption{Results of predictive Sensitivity values of HHEA-BNC for timeouts from 1,000s to 5,000s.}
    %
%
\end{table}%

\begin{table}[htbp] \scriptsize
  \centering
  \caption{Results of predictive Sensitivity values of HHEA-BNC for timeouts from 6,000s to 10,000s.}
    %
%
\end{table}%

\begin{table}[htbp] \scriptsize
  \centering
  \caption{Results of predictive GMean values of HHEA-BNC for timeouts from 1,000s to 5,000s.}
    %
%
\end{table}%

\begin{table}[htbp] \scriptsize
  \centering
  \caption{Results of predictive GMean values of HHEA-BNC for timeouts from 6,000s to 10,000s.}
    %
%
\end{table}%


\begin{table}[htbp] \scriptsize
  \centering
  \caption{Results of predictive Specificity values of Auto-WEKA-Bayes for timeouts from 1,000s to 5,000s.}
        %
%
\end{table}%

\begin{table}[htbp] \scriptsize
  \centering
  \caption{Results of predictive Specificity values of Auto-WEKA-Bayes for timeouts from 6,000s to 10,000s.}
    %
%
\end{table}%

\begin{table}[htbp] \scriptsize
  \centering
  \caption{Results of predictive Sensitivity values of Auto-WEKA-Bayes for timeouts from 1,000s to 5,000s.}
        %
%
\end{table}%

\begin{table}[htbp] \scriptsize
  \centering
  \caption{Results of predictive Sensitivity values of Auto-WEKA-Bayes for timeouts from 6,000s to 10,000s.}
    %
%
\end{table}%

\begin{table}[htbp] \scriptsize
  \centering
  \caption{Results of predictive GMean values of Auto-WEKA-Bayes for timeouts from 1,000s to 5,000s.}
    %
%
\end{table}%

\begin{table}[htbp] \scriptsize
  \centering
  \caption{Results of predictive GMean values of Auto-WEKA-Bayes for timeouts from 6,000s to 10,000s.}
    %
%
\end{table}%


\begin{table}[htbp] \scriptsize
  \centering
  \caption{Results of predictive Specificity values of Auto-WEKA-ALL for timeouts from 1,000s to 5,000s.}
    %
%
\end{table}%

\begin{table}[htbp] \scriptsize
  \centering
  \caption{Results of predictive Specificity values of Auto-WEKA-ALL for timeouts from 6,000s to 10,000s.}
    %
%
\end{table}%

\begin{table}[htbp] \scriptsize
  \centering
  \caption{Results of predictive Sensitivity values of Auto-WEKA-ALL for timeouts from 1,000s to 5,000s.}
    %
%
\end{table}%

\begin{table}[htbp] \scriptsize
  \centering
  \caption{Results of predictive Sensitivity values of Auto-WEKA-ALL for timeouts from 6,000s to 10,000s.}
    %
%
    \label{tab:anylabel}%
\end{table}%

\begin{table}[htbp] \scriptsize
  \centering
  \caption{Results of predictive GMean values of Auto-WEKA-ALL for timeouts from 1,000s to 5,000s.}
    %
%
\end{table}

\begin{table}[htbp] \scriptsize
  \centering
  \caption{Results of predictive GMean values of Auto-WEKA-ALL for timeouts from 6,000s to 10,000s.}
    %
%
\end{table}%

\begin{table}[htbp] \tiny
  \centering
  \caption{Results of predictive Specificity, Sensitivity and GMean values of AT across the 10 runtime limits.}
    %
%
\end{table}%

\begin{landscape}
\begin{table}[htbp] \footnotesize
  \centering
\caption{Results of predictive Specificity, Sensitivity and GMean values of PCC across the 10 runtime limits.}
    \begin{tabular}{lcccccccccccc}
    \toprule
          & \multicolumn{4}{c}{Specificity} & \multicolumn{4}{c}{Sensitivity} & \multicolumn{4}{c}{Gmean} \\
          \midrule
          & \multicolumn{1}{l}{1k} & \multicolumn{1}{l}{2k} & \multicolumn{1}{l}{3k-6k} & \multicolumn{1}{l}{7k-10k} & \multicolumn{1}{l}{1k} & \multicolumn{1}{l}{2k} & \multicolumn{1}{l}{3k-6k} & \multicolumn{1}{l}{7k-10k} & \multicolumn{1}{l}{1k} & \multicolumn{1}{l}{2k} & \multicolumn{1}{l}{3k-6k} & \multicolumn{1}{l}{7k-10k} \\
    \midrule
    CE    & 0.561 & 0.574 & 0.574 & 0.574 & 0.629 & 0.644 & 0.644 & 0.644 & 0.608 & 0.608 & 0.608 & 0.608 \\
    DM    & 0.550 & 0.550 & 0.550 & 0.550 & 0.675 & 0.675 & 0.675 & 0.675 & 0.607 & 0.607 & 0.607 & 0.607 \\
    MM    & 0.374 & 0.472 & 0.472 & 0.472 & 0.640 & 0.674 & 0.674 & 0.674 & 0.557 & 0.557 & 0.557 & 0.557 \\
    SC    & 0.351 & 0.351 & 0.351 & 0.351 & 0.778 & 0.778 & 0.778 & 0.778 & 0.507 & 0.507 & 0.507 & 0.507 \\
    DNA3  & 0.718 & 0.718 & 0.718 & 0.718 & 0.879 & 0.879 & 0.879 & 0.879 & 0.790 & 0.790 & 0.790 & 0.790 \\
    DNA11 & 0.448 & 0.448 & 0.448 & 0.448 & 0.749 & 0.749 & 0.749 & 0.749 & 0.571 & 0.571 & 0.571 & 0.571 \\
\bottomrule
    \end{tabular}%
  \label{tab:addlabel}%
\end{table}%
\end{landscape}

\small

\bibliographystyle{apalike}
\bibliography{paper}


\end{document}